\documentclass[lettersize,journal]{IEEEtran}
\usepackage{amsmath,amsfonts}
\usepackage{algorithmic}
\usepackage{algorithm}
\usepackage{array}
\usepackage[caption=false,font=normalsize,labelfont=sf,textfont=sf]{subfig}
\usepackage{textcomp}
\usepackage{stfloats}
\usepackage{url}
\usepackage{verbatim}
\usepackage{graphicx}
\usepackage{cite}
\usepackage{multirow}
\usepackage{tabularx} 
\usepackage{xcolor}
\usepackage{makecell}
\usepackage{pifont}

\newcommand{\Cov}{\mathrm{Conv}}
\newcommand{\SN}{\mathrm{SN}}
\newcommand{\tdBN}{\mathrm{tdBN}}
\newcommand{\MP}{\mathrm{MPool}}

\begin{document}

\makeatletter
\long\def\@makecaption#1#2{%
\ifx\@captype\@IEEEtablestring%
\footnotesize\bgroup\par\centering\@IEEEtabletopskipstrut%
{\normalfont\footnotesize\bfseries #1: \normalfont\footnotesize #2}%
\par\addvspace{0.5\baselineskip}\egroup%
\@IEEEtablecaptionsepspace
\else
\@IEEEfigurecaptionsepspace
\setbox\@tempboxa\hbox{\normalfont\footnotesize {#1.}\nobreakspace\nobreakspace #2}%
\ifdim \wd\@tempboxa >\hsize%
\setbox\@tempboxa\hbox{\normalfont\footnotesize {#1.}\nobreakspace\nobreakspace}%
\parbox[t]{\hsize}{\normalfont\footnotesize\noindent\unhbox\@tempboxa#2}%
\else%
\hbox to\hsize{\normalfont\footnotesize\box\@tempboxa\hfil}%
\fi\fi}
\makeatother

\title{SpikeDet: Better Firing Patterns for Accurate and Energy-Efficient Object Detection with Spiking Neural Networks}

\author{Yimeng Fan,
Changsong Liu,
Mingyang Li,
Dongze Liu,
Yuting Su,
Yanyan Liu$^{\dagger}$,
and 
Wei Zhang$^{\dagger}$\thanks{$\dagger$ Corresponding author: Yanyan Liu, Wei Zhang.}
\thanks{Y. Fan, C. Liu, M. Li, D. Liu, and W. Zhang are with the School of Microelectronics, and Y. Su is with the School of Electrical and Information Engineering, Tianjin University, Tianjin 300072, China. E-mail: \{yimengfan, changsong, limingyang97, ldz, tjuzhangwei, ytsu\}@tju.edu.cn.}
\thanks{Y. Liu is with the Optoelectronic Thin Film Device and Technology Research Institute, Nankai University, Tianjin 300350, China. E-mail: lyytianjin@nankai.edu.cn.}}



\maketitle

\begin{abstract}
Spiking Neural Networks (SNNs) are the third generation of neural networks. They have gained widespread attention in object detection due to their low energy consumption and biological interpretability.
However, existing SNN-based object detection methods suffer from local firing saturation, where adjacent neurons concurrently reach maximum firing rates, especially in object-centric regions. This abnormal neuron firing pattern reduces the feature discrimination capability and detection accuracy, while also increasing the firing rates that prevent SNNs from achieving their potential energy efficiency.
To address this problem, we propose SpikeDet, a novel spiking object detector that optimizes firing patterns for accurate and energy-efficient detection. 
Specifically, we design a spiking backbone network, MDSNet, which effectively adjusts the membrane synaptic input distribution at each layer, achieving better neuron firing patterns during spiking feature extraction.
For the neck, to better utilize and preserve these high-quality backbone features, we introduce the Spiking Multi-direction Fusion Module (SMFM), which realizes multi-direction fusion of spiking features, enhancing the multi-scale detection capability of the model.
Furthermore, we propose the Local Firing Saturation Index (LFSI) to quantitatively measure local firing saturation.
Experimental results validate the effectiveness of our method.
On the COCO 2017 dataset, it achieves 52.2\% AP, outperforming previous SNN-based methods by 3.3\% AP while requiring only half the energy consumption. On object detection sub-tasks, including event-based GEN1, underwater URPC 2019, low-light ExDARK, and dense scene CrowdHuman datasets, SpikeDet also achieves the best performance. 
\end{abstract}


\section{Introduction}\label{intro}
\IEEEPARstart{S}{piking} Neural Networks (SNNs) have emerged as a promising alternative to traditional Artificial Neural Networks (ANNs) \cite{ren2026language}.
Unlike ANNs that rely on continuous values, spiking neurons in SNNs emulate biological neurons, computing and communicating through discrete spiking signals \cite{maass_networks_1997}.
This spike-driven mechanism can avoid the computational burden of multiplication operations, instead relying on simpler addition operations, providing significant potential for energy-efficient computing \cite{chen2025retain}.
Furthermore, the temporal dynamics of these spikes provide SNNs with distinct representation capabilities \cite{yao_spikedriven_2024}.
Motivated by these characteristics, SNNs have been increasingly applied to various computer vision tasks \cite{xie2024eisnet, cordone_object_2022, barchid2023spiking, xiao2026learning}.

Among these, SNNs have achieved competitive performance with ANNs on simpler tasks such as image classification \cite{yao_spikedriven_2024, yao_scaling_2025}.
However, for more complex tasks like object detection, which require simultaneous identification of categories, locations, and sizes \cite{du2025enhanced, zhang2026plvam}, SNNs still face significant challenges.
Specifically, performance gaps persist between SNN-based and ANN-based detectors \cite{kim_spikingyolo_2020, cordone_object_2022, su_deep_2023, fan_sfod_2024, luo_integervalued_2025, yao_scaling_2025}.
More critically, the energy efficiency advantages of SNNs diminish as performance improves \cite{luo_integervalued_2025, yao_scaling_2025}.

\begin{figure}[!t]
\centering
    \includegraphics[width=1.\columnwidth]{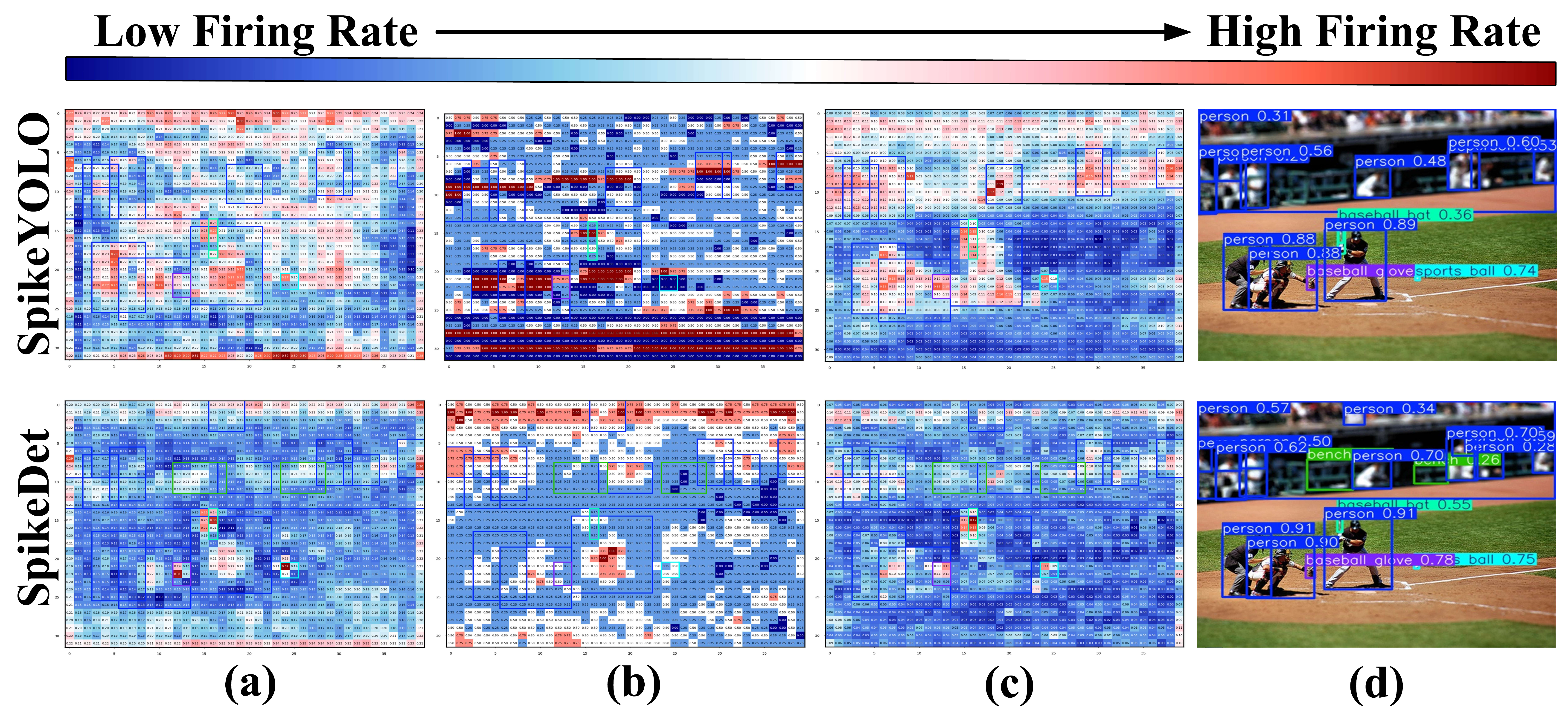}
\caption{\textbf{Visualization of local firing saturation problem in SNN-based object detector on COCO dataset.}
Each pixel represents neuron firing rate.
(a) and (b) show pre-detection-head feature maps, which determine both classification and regression.
(a) averages the $4D$ spike tensor $([T, C, H, W])$ across time and channel dimensions to show overall spatial firing distribution, while (b) selects a representative channel and averages across time to reveal individual neuron firing patterns.
(c) shows neuron firing patterns of backbone features at $1/16$ resolution.
(d) shows final detection results.}
\label{fig_1}
\vspace{-0.4cm}
\end{figure}

These limitations motivate us to identify a fundamental problem in existing SNN-based detectors. 
As shown in Fig. \ref{fig_1}, we observe that spiking neurons consistently reach maximum firing rates (firing saturation) in certain spatial regions, particularly in central regions of bounding boxes where object information is most concentrated \cite{zhang_bridging_2020}.
\textbf{We define this phenomenon, where multiple spatially adjacent neurons concurrently exhibit saturated firing patterns, as local firing saturation.}
These saturated neurons are detrimental for SNN-based object detection. In SNNs, features are encoded through neuron firing patterns. 
When multiple adjacent neurons reach firing saturation, they produce identical representations, reducing feature discriminability. 
This issue affects both components of object detection, consequently impacting detection accuracy. 
For classification, reduced feature discriminability causes multiple anchor points to produce similar confidence scores, preventing concentrated object representation and degrading overall confidence. For regression, this generates redundant bounding boxes that waste anchor resources and cause missed detections in overlapping scenes, such as the rear person and bench in Fig. \ref{fig_1} (d).
Additionally, this local firing saturation issue increases the firing rate of SNN detectors, consequently raising their energy consumption.

\begin{figure}[t]
\centering
\includegraphics[width=.95\columnwidth]{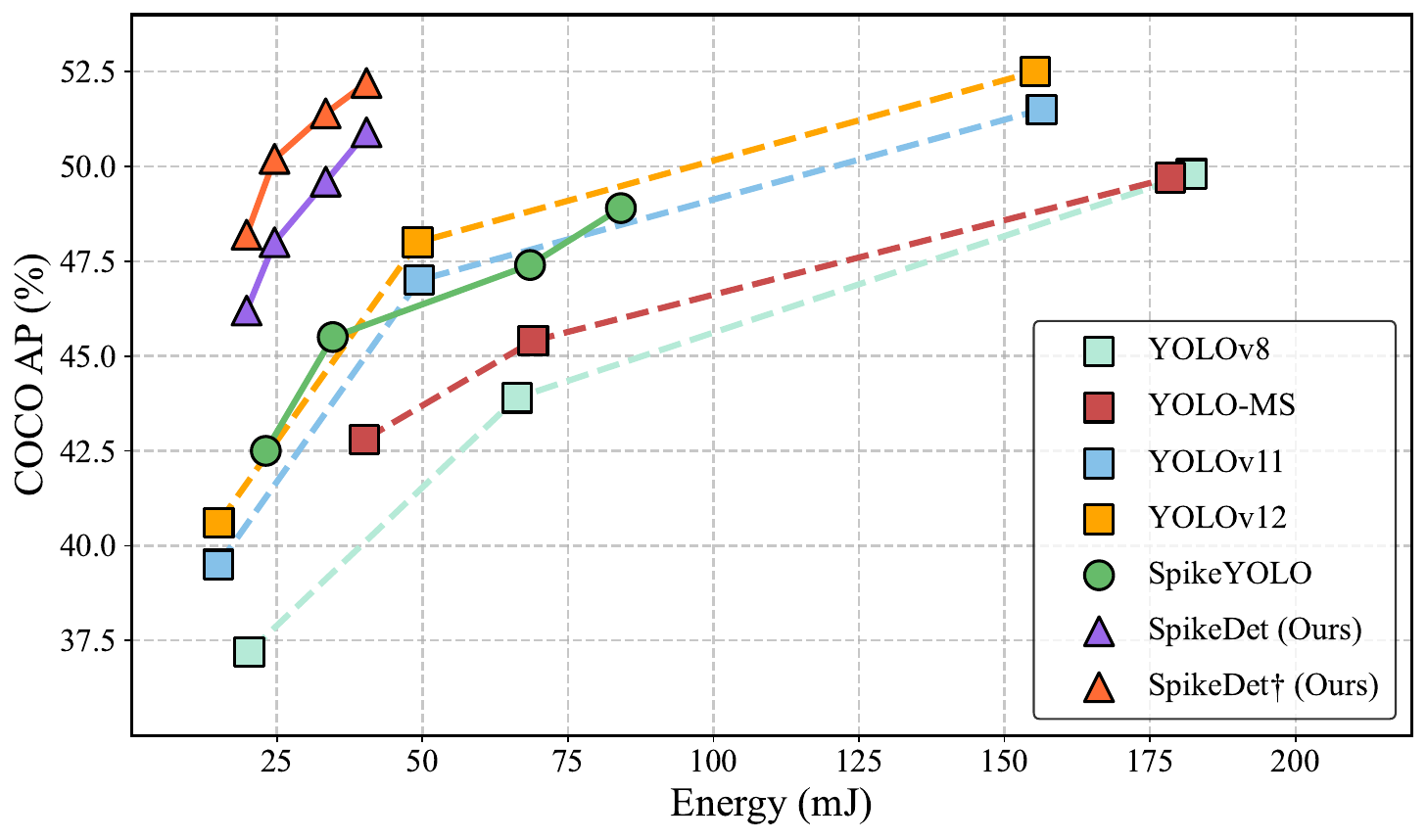}
\caption{\textbf{Comparisons with other state-of-the-art methods in COCO AP and Energy consumption on the COCO 2017 dataset.} Squares represent ANN-based object detectors, circles represent SNN-based object detectors, and triangles represent our methods.} 
\label{fig_2}
\vspace{-0.4cm}
\end{figure}

To address this issue, we propose SpikeDet, a novel spiking object detector. By optimizing neuron firing patterns, SpikeDet mitigates local firing saturation and achieves accurate, energy-efficient object detection.
The core innovation of our approach is the introduction of a spiking backbone network, \textbf{M}embrane-based \textbf{D}eformed \textbf{S}hortcut Residual \textbf{Net}work (MDSNet).
Current high-accuracy SNN object detectors \cite{yao_spikedriven_2024, su_deep_2023, luo_integervalued_2025, yao_scaling_2025} employ membrane-based shortcuts \cite{hu_advancing_2024} to achieve residual learning in their backbone networks for enhancing feature extraction capacity. 
However, this results in the residual path output being directly added to the shortcut path at each layer.
As we theoretically analyze in Section \ref{mdsnet_motivation}, such repeated path summation causes membrane synaptic input variance to accumulate across successive layers.
This variance amplification significantly increases the probability of neurons receiving excessively large synaptic inputs.
Consequently, neurons more frequently reach firing saturation. Combined with the inherently high neural activity in information-concentrated regions, this causes adjacent neurons to concurrently saturate, leading to the local firing saturation problem shown in Fig. \ref{fig_1}(c).
For this, we propose the Membrane-based Deformed Shortcut (MDS), which enables the identity mapping to stabilize its output distribution, fundamentally mitigating the local firing saturation problem.
Based on MDS and residual architectures, we design MDSNet, which effectively combines the advantages of both, delivering powerful feature extraction capacity for SNN-based object detectors.

Multi-scale feature fusion also plays an important role in mitigating local firing saturation. For this, we propose the Spiking Multi-direction Fusion Module (SMFM).
In traditional spiking fusion methods \cite{fan_sfod_2024, luo_integervalued_2025}, information flows along limited paths. 
Such insufficient fusion raises two issues. 
First, features from different scales may simultaneously activate at object centers. Limited fusion prevents their effective integration, causing their responses to accumulate and leading to local firing saturation.
Second, with limited fusion, shallow detail-sensitive features lack sufficient semantic modulation when integrated with deeper features, causing local firing saturation at regions with high local contrast in the fused representations.
Multi-direction fusion creates multiple pathways, allowing features to undergo repeated refinement. This allows better integration of information across scales, mitigating local firing saturation while enhancing multi-scale feature utilization.
Additionally, to quantitatively evaluate the capability of our method in addressing local firing saturation, we propose the Local Firing Saturation Index (LFSI). 
This metric measures the proportion of local firing saturation occurrences among all spatial neighborhoods in the network, reflecting its severity.

In summary, the contributions of this work are as follows:

(1) We propose SpikeDet, an accurate and energy-efficient spiking object detector.
SpikeDet achieves better neuron firing patterns by optimizing the backbone and neck networks, thereby mitigating the local firing saturation problem faced by SNN-based object detectors. 

(2) For the backbone, we propose MDSNet, which integrates MDS to regulate the membrane synaptic input distribution of subsequent neurons, enhancing firing pattern stability. 
For the neck, we design SMFM that implements multi-direction feature fusion in SNNs, preserving stable firing patterns while improving the model's multi-scale detection capability.

(3) As demonstrated in Fig. \ref{fig_2}, our method achieves superior performance while maintaining the lowest energy consumption on the COCO 2017 dataset \cite{lin_microsoft_2014}. Moreover, the proposed LFSI quantitatively verifies its effectiveness in addressing local firing saturation.


\section{Related Work}
\subsection{Spiking Neural Networks}
SNNs are designed to mimic biological behavior more accurately than ANNs through spiking neurons \cite{yao_spikedriven_2024}. However, the non-differentiable nature of neuron spike firing prevents SNNs from being trained with traditional backpropagation. To address this, researchers have proposed two approaches, ANN2SNN conversion and direct training. 
The former requires thousands of time steps and is suited only for static datasets \cite{kim_spikingyolo_2020}. The latter leverages surrogate gradients \cite{fan_sfod_2024}, achieving strong performance within limited time steps. Therefore, we employ the direct training strategy in this study.
To further advance SNN performance and realize deeper architectures, researchers have proposed numerous SNN-based residual learning methods \cite{zheng_going_2021, fang_deep_2021, hu_advancing_2024}. Among them, SEW-ResNet \cite{fang_deep_2021} and MS-ResNet \cite{hu_advancing_2024} address gradient issues and train networks exceeding 100 layers. However, the former introduces non-spiking convolutions caused by spike addition. The latter solves this issue but introduces a local firing saturation problem that we identify in this work. This paper further investigates SNN characteristics to enhance feature extraction while maintaining a full-spiking network.

Intuitively, our work is related to SNN-based attention mechanisms \cite{yao2023attention, yao2023inherent}, as both aim to improve spiking feature representation. 
However, there are fundamental differences. 
Firstly, in terms of problem nature, \cite{yao2023attention, yao2023inherent} focus on identifying and suppressing redundant spike firing in SNNs, which is essentially a feature-level selection problem. In contrast, our work addresses local firing saturation in SNN-based object detection. The former concerns excessive firing of uninformative spikes, while the latter involves identical firing patterns among adjacent neurons, leading to a loss of feature discriminability.
Secondly, the two works differ in methodology. \cite{yao2023attention, yao2023inherent} propose plug-and-play attention modules. In contrast, our method addresses local firing saturation through structural improvements at the network level, namely MDS in the backbone and SMFM in the neck.

\subsection{Object Detection}
Object detection is an important and highly complex task in computer vision. 
Early object detection relied on hand-crafted features and sliding window approaches, achieving remarkable accuracy in face detection \cite{garcia1999face}. 
With the development of ANNs, ANN-based methods gradually became mainstream. These can be categorized into two-stage and one-stage object detectors.
Two-stage methods consist of a proposal generator and a region-wise prediction subnetwork \cite{chen2023ddod}. Although they can achieve good performance, they have gradually been replaced by one-stage methods due to their complex design and inefficient inference. 
One-stage methods directly generate object locations and classifications, significantly reducing computational complexity while maintaining competitive accuracy. Representative examples include YOLO \cite{yolov8_ultralytics, tian_yolov12_2025} and DETR \cite{carion_endtoend_2020} series.

\subsection{Object Detection with Spiking Neural Networks}
Early attempts to apply SNNs to object detection using ANN2SNN conversion \cite{kim_spikingyolo_2020} face the challenges of high latency, poor performance, and incompatibility with event cameras.
VC-DenseNet \cite{cordone_object_2022} and EMS-YOLO \cite{su_deep_2023} pioneer direct training for SNN-based object detection, achieving promising results on both static and event-based datasets. 
SFOD \cite{fan_sfod_2024} first introduces an efficient Spiking Fusion Module for SNNs. SpikeYOLO \cite{luo_integervalued_2025} proposes the I-LIF neuron to address the impact of quantization errors from spiking neurons on object detection, significantly improving the performance of SNN-based detectors. 
However, both approaches overlook the fundamental differences between SNN and ANN-based feature representations: SNN features are represented through neuron firing patterns. This oversight amplifies their performance gap with ANN-based detectors. Hence, our research enhances the SNN detector by improving neuron firing patterns.

\begin{figure*}[t]
\centering
\includegraphics[width=.95\textwidth]{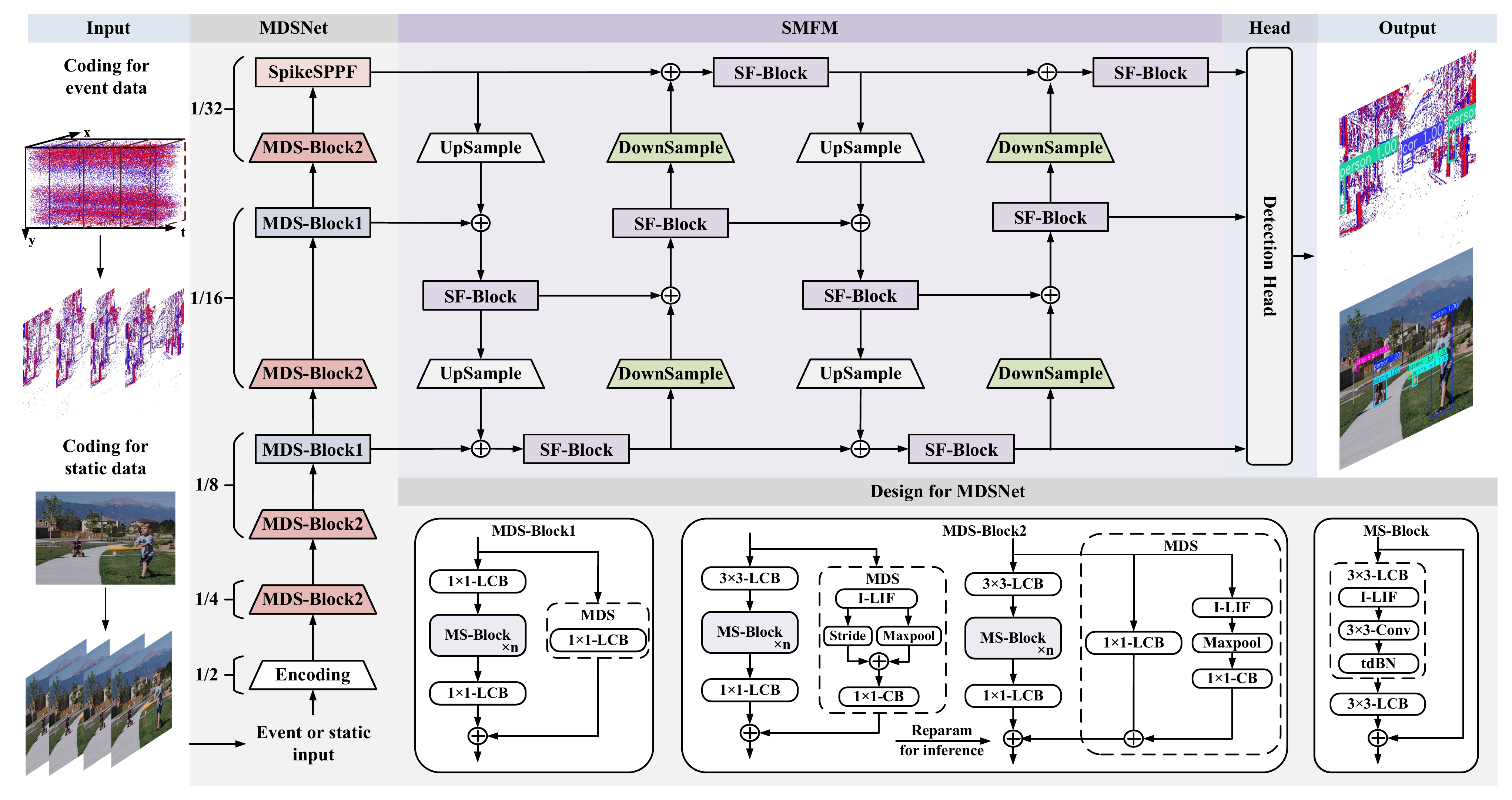}
\caption{\textbf{The architecture of SpikeDet.} 
SpikeDet comprises MDSNet, SMFM, and the SpikeYOLO Detection Head \cite{luo_integervalued_2025}.
The model receives two types of inputs: event and static data, with the input coding and output represented in the figure.
The core of SpikeDet is MDSNet, which consists of 5 stages. The downsampling factor for each stage is indicated in the figure, where MDS-Block1 and MDS-Block2 perform feature extraction without and with downsampling, respectively.
By incorporating MDS, this architecture successfully stabilizes the firing patterns at each stage, alleviating the local firing saturation problem. 
Additionally, SMFM enables multi-direction feature fusion, allowing features to undergo multiple refinements, thereby improving the model's capability to detect multi-scale objects and preserving neuron firing patterns.
Note that 1$\times$1-CB denotes 1$\times$1-LCB without the I-LIF neuron. 
}
\label{fig_3}
\vspace{-0.36cm}
\end{figure*}

\section{Methodology}  
\subsection{Overview}
The architecture of our proposed SpikeDet is shown in Fig. \ref{fig_3}. 
As illustrated, the model adopts the widely used backbone-neck-head object detection architecture \cite{fan_sfod_2024, chen_yoloms_2025, du2025enhanced}. 
For the backbone, we propose MDSNet to extract multi-scale features with more stable firing patterns.
For the neck, we design a Spiking Multi-direction Fusion Module (SMFM), which enables the model to enhance backbone features through multi-direction fusion.
For the detection head, which primarily generates detection results from formed features, establishing firing pattern stability is not critical.
Therefore, we directly adopt the existing SpikeYOLO Detection Head \cite{luo_integervalued_2025}.

The spike-driven characteristics of SNNs make them naturally suited to event data. Consequently, SNN-based detectors are now explored on both static and event-based datasets.
To handle both data types effectively and exploit the temporal dynamics of SNNs, we encode inputs as $4D$ tensors $[T, C, H, W]$, denoting time steps, channels, height, and width.
For static data, we employ the direct coding method \cite{yao_spikedriven_2024}, replicating the input $T$ times to match this format.
For event data, each event is defined as $e_k = (t_k, p_k, x_k, y_k)$, where $(x_k, y_k)$ are spatial coordinates, $t_k$ is the timestamp, and $p_k$ is the polarity. Accordingly, we adopt the coding approach from \cite{hu_advancing_2024, luo_integervalued_2025}. This approach selects a time window, divides it into $T$ temporal bins, and integrates events within each bin into a frame, forming the 4D tensor above.
Finally, we use Rate Decoding \cite{fan_sfod_2024} for model output decoding, followed by Non-Maximum Suppression (NMS) post-processing.

\subsection{The Preliminaries of Spiking Neurons}
To balance biological plausibility and computational complexity in SNNs, various spiking neuron models have been proposed, including the Hodgkin-Huxley \cite{hodgkin_quantitative_1952}, Integrate-and-Fire (IF) \cite{gerstner2014neuronal}, Leaky Integrate-and-Fire (LIF) \cite{abbott_lapicques_1999a}, and Integer Leaky Integrate-and-Fire (I-LIF) \cite{luo_integervalued_2025} models, etc.
In this work, we employ the I-LIF neuron due to its effectiveness in reducing the quantization error of SNNs.
The I-LIF neuron model is represented by:
\begin{equation}\label{ILIF1}
\mathbf{u}^{t,n} = \tau \mathbf{h}^{t-1,n} + \mathbf{x}^{t,n},
\end{equation}
\begin{equation}\label{ILIF2}
\mathbf{o}^{t,n} = \mathit{Clip} \left( \mathit{round}(\mathbf{u}^{t,n}), 0, D \right),
\end{equation}
\begin{equation}\label{ILIF3}
\mathbf{h}^{t,n} = \mathbf{u}^{t,n} - V_{th} \cdot \mathbf{o}^{t,n}.
\end{equation}
Here, $\mathbf{u}$ represents the membrane potential, $\mathbf{h}$ the membrane potential retained from the previous time step, $\mathbf{x}$ the pre-synaptic input, and $\mathbf{o}$ the output spike. Variables $t$ and $n$ denote time step and layer number, respectively. 
Function $\mathit{Clip} \left(x, min, max \right)$ constrains $x$ within $[min, max]$, while $\mathit{round}(x)$ rounds $x$ to the nearest integer. $D$ is the upper limit hyperparameter for integer-valued spike activation, $\tau$ is the membrane potential decay constant, and $V_{th}$ is the membrane potential threshold.
The core of the I-LIF neuron lies in Eq. \ref{ILIF2}, which enables neurons to generate integer spikes during training and can convert them to the sum of binary spikes in $D$ time steps of Soft-Reset IF neurons during inference \cite{yao_scaling_2025}. 

\subsection{MDSNet}\label{mdsnet}
\subsubsection{Motivation}\label{mdsnet_motivation}
Object detection is a complex task that demands strong feature extraction capabilities of the backbone network. 
In SNNs, to address this requirement, extensive studies have been conducted on SNN-friendly residual learning approaches \cite{hu_advancing_2024, fang_deep_2021, zheng_going_2021}, as constructing deep networks through residual learning has been proven beneficial in ANNs.
Among these efforts, MS-ResNet \cite{hu_advancing_2024} is the most successful approach, featuring its membrane-based shortcut. This design enables efficient information propagation in deep SNN-based models.
Building upon this, subsequent backbones such as EMS-ResNet \cite{su_deep_2023} and S-Backbone \cite{luo_integervalued_2025} have been proposed, enabling the development of advanced SNN-based object detectors.
Taking EMS-ResNet as an example, this network can be described as follows,
\begin{equation}
\psi = \tdBN \circ \Cov \circ \SN,
\end{equation}
\begin{equation}\label{ms-block}
\mathbf{y}_{I}^{t,l}=\psi\left(\psi(\mathbf{x}^{t,l})\right) + \mathbf{x}^{t,l},
\end{equation}
\begin{equation}\label{ems-block1}
\scalebox{0.97}{$
\mathbf{y}_{D1}^{t,l}=\psi\left(\psi(\mathbf{x}^{t,l})\right)+\mathrm{Cat}[\MP(\mathbf{x}^{t,l}),\psi(\MP(\mathbf{x}^{t,l}))],
$}
\end{equation}
\begin{equation}\label{ems-block2}
\mathbf{y}_{D2}^{t,l}=\psi\left(\psi(\mathbf{x}^{t,l})\right)+\psi(\MP(\mathbf{x}^{t,l})).
\end{equation}
Here, \( \mathbf{y} \) represents the output and \( \mathbf{x} \) represents the input. \( \tdBN \), \( \Cov \), \( \SN \), \( \MP \), and \( \mathrm{Cat} \) denote threshold-dependent Batch Normalization proposed in \cite{zheng_going_2021}, convolution, I-LIF neurons, Maxpool, and channel dimension concatenation, respectively. $\psi$ denotes I-LIF-Conv-tdBN (LCB).
The superscript \( l \) denotes the \( l \)-th residual block, while subscripts \( D \) and \( I \) indicate whether the residual block performs downsampling or not. Accordingly, Eq. \ref{ms-block} represents the MS-Block without downsampling, while Eqs. \ref{ems-block1} and \ref{ems-block2} represent the downsampling blocks EMS-Block1 and EMS-Block2, respectively.

When the model goes deeper, MS-Blocks are stacked repeatedly. Let $\mathbf{y}_R^{t,l}$ and $\mathbf{y}_S^{t,l}$ denote the residual and shortcut path outputs of a block. The block output variance $\text{Var}[\mathbf{y}_I^{t,l}] = \text{Var}[\mathbf{y}_R^{t,l} + \mathbf{y}_S^{t,l}]$ can be expanded as
\begin{equation}
\text{Var}[\mathbf{y}_I^{t,l}] = \text{Var}[\mathbf{y}_R^{t,l}] + 2\text{Cov}[\mathbf{y}_R^{t,l}, \mathbf{y}_S^{t,l}] + \text{Var}[\mathbf{y}_S^{t,l}].
\end{equation}
To further analyze this, we derive the following proposition. 

\noindent\textbf{Proposition 1.}
\textit{For $d$ stacked LCB layers ($d \geq 1$) with arbitrary kernel sizes, under the convolution zero-mean weight and zero bias initialization, the output at the $(n+d)$-th layer $\mathbf{y}^{t,n+d}$ is uncorrelated with the input at the $n$-th layer $\mathbf{x}^{t,n}$.}

\noindent\textbf{Proof} The detailed proof is presented in Appendix A-A.

Based on Proposition 1, the residual path $\mathbf{y}_R^{t,l} = \psi(\psi(\mathbf{x}^{t,l}))$ is the output of 2 stacked LCB layers processing the input $\mathbf{x}^{t,l}$, while the shortcut path $\mathbf{y}_S^{t,l} = \mathbf{x}^{t,l}$. Therefore, we have $\text{Cov}[\mathbf{y}_R^{t,l}, \mathbf{y}_S^{t,l}] = 0$.
Thus 
$\text{Var}[\mathbf{y}_I^{t,l}] = \text{Var}[\mathbf{y}_R^{t,l}] + \text{Var}[\mathbf{y}_S^{t,l}]$.
For $k$ stacked MS-Blocks, we can recursively derive that 
\vspace{-0.2cm}
\begin{equation}\label{var_sum}
\text{Var}[\mathbf{y}_I^{t,k}] = \text{Var}[\mathbf{y}_S^{t,1}] + \sum_{l=1}^{k}{\text{Var}[\mathbf{y}_R^{t,l}]}.  
\vspace{-0.2cm}
\end{equation}
Therefore, the variance of \( \mathbf{y}_I^{t,l} \) becomes increasingly large as the network deepens.
This results in unstable membrane synaptic input distribution in subsequent neurons.

\begin{figure}[!t]
\centering
    \includegraphics[width=.99\columnwidth]{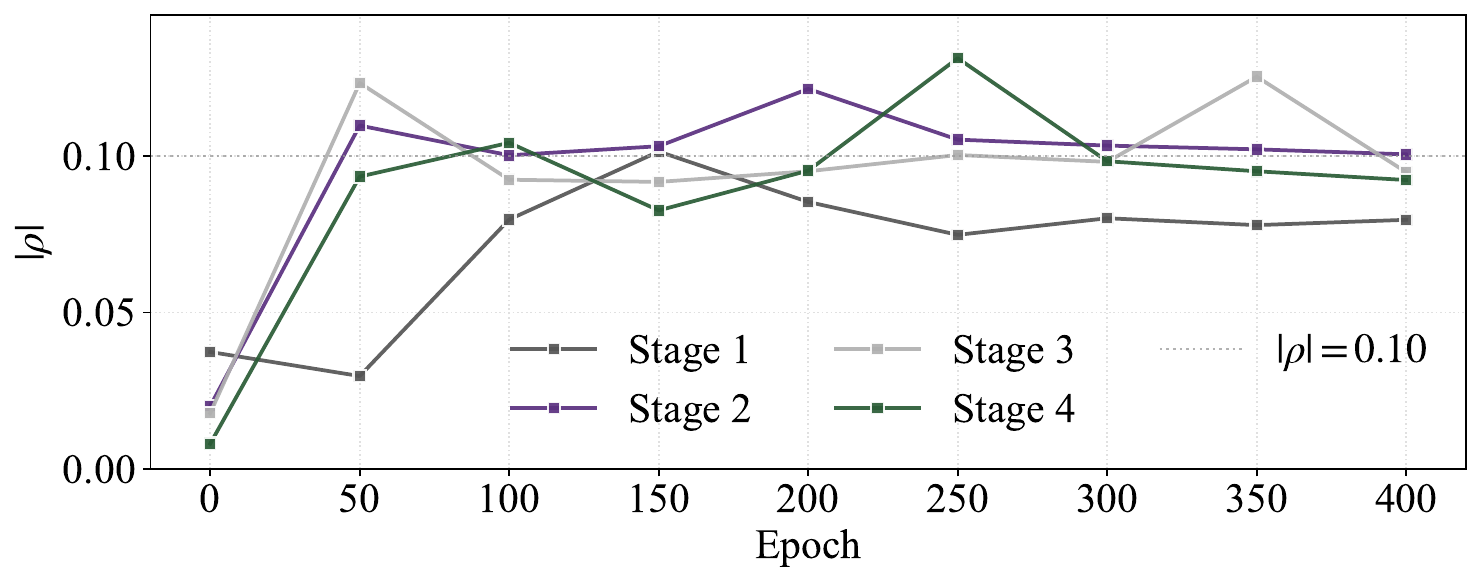}
\caption{\textbf{Experimental measurement of $|\rho(\mathbf{y}_R^{t,l}, \mathbf{y}_S^{t,l})|$ across training epochs in EMS-ResNet34 on COCO.} 
The reported value for each stage is averaged over all MS-Blocks within that stage.
}
\label{fig_rho}
\vspace{-0.3cm}
\end{figure}

This property holds under standard initialization schemes \cite{he2015delving}. It continues to hold approximately during and after training due to weight decay and the stabilizing effect of tdBN. 
As shown in Fig.~\ref{fig_rho}, we measure the correlation coefficient 
$|\rho(\mathbf{y}_R^{t,l}, \mathbf{y}_S^{t,l})| = \left|\frac{\text{Cov}[\mathbf{y}_R^{t,l}, \mathbf{y}_S^{t,l}]}{\text{std}[\mathbf{y}_R^{t,l}]\cdot\text{std}[\mathbf{y}_S^{t,l}]}\right|$ on EMS-ResNet34. Compared to directly measuring $\text{Cov}$, the calculation of correlation coefficient eliminates the influence of activation scale, enabling consistent comparison across stages. 
$|\rho|$ remains below 0.04 at initialization and stabilizes around 0.10 throughout training.
By the AM-GM inequality, the relative error induced by dropping the covariance term satisfies
\begin{equation}
\scalebox{0.85}{$\displaystyle
\frac{2\text{Cov}\left[\mathbf{y}_R^{t,l},\mathbf{y}_S^{t,l}\right]}{\text{Var}\left[\mathbf{y}_R^{t,l}\right]+\text{Var}\left[\mathbf{y}_S^{t,l}\right]}
=\frac{2\rho\cdot \text{std} \left[\mathbf{y}_R^{t,l}\right]\cdot \text{std} \left[\mathbf{y}_S^{t,l}\right]}{\text{Var}\left[\mathbf{y}_R^{t,l}\right]+\text{Var}\left[\mathbf{y}_S^{t,l}\right]}
\le\rho.$}
\end{equation}
Therefore, $|\rho|$ remaining around $0.10$ throughout training implies a relative error of within 10\%, experimentally confirming the validity of the above approximation.

\begin{figure*}[t]
\centering
\includegraphics[width=.95\textwidth]{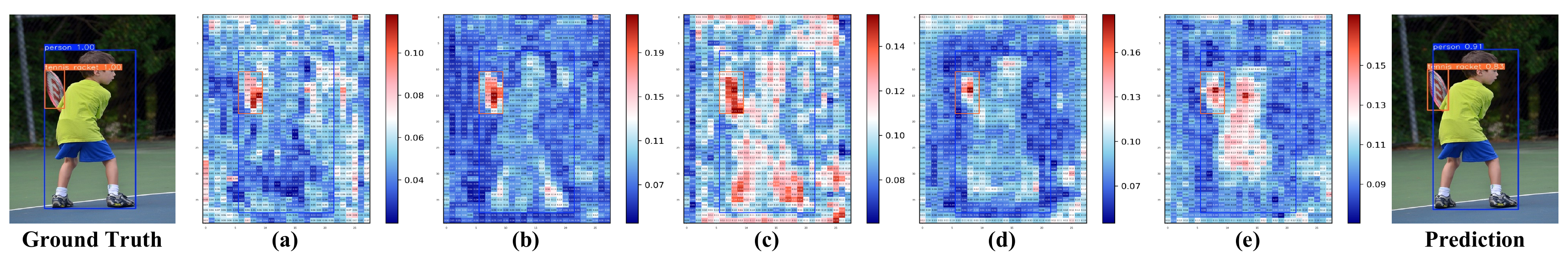}
\caption{\textbf{Influence of multi-direction feature fusion on firing patterns of SNN-based detector.} 
We visualize feature maps at the 1/16 downsampling stage, averaging across $T$ and $C$ dimensions to reveal overall firing patterns.
Figures (a) to (e) show neuron firing patterns for feature maps with no fusion, one-way, two-way, three-way, and four-way fusion, respectively.} 
\label{fig_5}
\vspace{-0.4cm}
\end{figure*}

\noindent\textbf{Proposition 2.}
\textit{For I-LIF neurons, the variance of the membrane potential is proportional to the variance of the input, i.e., $\text{Var}[\mathbf{u}^{t,n}] \propto \text{Var}[\mathbf{x}^{t,n}]$.}

\noindent\textbf{Proof} The detailed proof is presented in Appendix A-B.

With Proposition 2, we can conclude that the aforementioned variance accumulation in synaptic input directly creates unstable membrane potential distributions. 
According to Eq. \ref{ILIF2}, this instability increases the probability of neurons firing large integer-valued spikes, thereby exacerbating the local firing saturation problem. This severely impacts object detection tasks as discussed in Section \ref{intro}.

\subsubsection{MDSNet Architecture}
To address these issues, as shown in Fig. \ref{fig_3}, we propose MDSNet, a novel SNN-based backbone network.
On the shortcut path, we propose a Membrane-based Deformed Shortcut (MDS), which incorporates 1$\times$1-LCB, i.e., LCB using 1$\times$1 convolution. This structure converts input to spikes through the I-LIF layer and processes them through a 1$\times$1 convolution, which improves expressive capability. The tdBN component further provides this path with the ability to adjust the output distribution. 
As a result, the output variance $\text{Var}[\mathbf{y}_{MDS}^{t,l}]$ equals a constant $c$, independent of the input variance. 
This fundamentally breaks the cross-block variance accumulation in Eq. \ref{var_sum}, which is also experimentally verified in Section \ref{exp_mds}.
As analyzed in the previous section, MDS reduces the probability of neurons firing saturated spikes. Consequently, even in regions where neural activity is inherently high, adjacent neurons are less likely to reach saturation concurrently, thereby mitigating the local firing saturation problem.

On the residual path, we place MS-Blocks in it rather than on the main path. It provides a clean gradient propagation path for deep networks. Additionally, to avoid the problems with stacked MS-Blocks, we place a 1$\times$1-LCB at both the input and output to control variance. 
Through improvements on both paths, we ensure the output distribution stability of each block in MDSNet. Meanwhile, structural enhancements preserve the efficient information propagation advantage offered by the membrane-based shortcut in deep networks.

Specifically, MDSNet consists of MDS-Block1 and MDS-Block2. MDS-Block1 is designed without downsampling, following the structure described above. 
MDS-Block2 serves as the downsampling block. Its residual path is identical to MDS-Block1. 
In the shortcut path, we modify MDS to support downsampling. Maxpool and fixed stride downsampling are employed in parallel within MDS. Maxpool selects the most salient features, while fixed stride downsampling preserves spatial distribution information. 
The two paths complement each other and improve the information-preserving ability of MDS during downsampling. 
As shown in Fig. \ref{fig_3}, we position them after the I-LIF layer. 
Compared to placing them before I-LIF, this allows both paths to operate on bounded spikes without amplifying the I-LIF input. As a result, the probability of neuron firing saturation is reduced, alleviating the local firing saturation problem.
Furthermore, at inference time, we apply reparameterization \cite{ding2021repvgg} to split it into two separate paths. Each path output is within $[0, D]$, allowing I-LIF to correctly convert integer values into spike trains \cite{yao_scaling_2025} and preserving the spike-driven computation mechanism of the model.

\subsection{SMFM}\label{SMFM}
Research in feature fusion for SNNs is limited to one-way and two-way fusion \cite{fan_sfod_2024, luo_integervalued_2025}, where large-scale and small-scale features are combined only a limited number of times. As discussed in Section \ref{intro}, such insufficient fusion leads to local firing saturation in both object centers and regions with high local contrast. This makes multi-directional feature fusion more critical in SNNs than in ANNs.
To address these issues, we propose the Spiking Multi-direction Fusion Module (SMFM). It enables the model to repeatedly integrate multi-scale features through iterative fusion, allowing more comprehensive feature fusion.
Fig. \ref{fig_5} provides visualization evidence that multi-direction fusion effectively mitigates local firing saturation.

The structure of the SMFM is shown in Fig. \ref{fig_3}. In this module, feature maps ranging from 1/8 to 1/32 are fused. 
During the fusion process, we first fuse feature maps of lower resolution with those of higher resolution through upsampling. Following this, we employ downsampling to further integrate these fused feature maps. This process is then repeated iteratively.
Next, we will discuss the methods for upsampling, downsampling, and spiking feature fusion.

To achieve upsampling/downsampling, we use Nearest-Neighbor Interpolation (NNI) with 1$\times$1-LCB and 3$\times$3-LCB with a stride of 2 as core elements.
For feature fusion, we multiply the feature maps by learnable constants and then add them together. 
These constants can be fused with the previous tdBN layer. As the features serve as presynaptic inputs to neurons, this method is compatible with SNNs.
In this way, we achieve better feature alignment through adaptive weighting of different-scale features without extra parameters.
Then, we propose the Spiking Fusion Block (SF-Block) to enhance the fused features.
This block consists of two sub-blocks. The first has two paths, with one containing two stacked 3$\times$3-LCBs and another containing a 3$\times$3-LDCB (with internal depthwise convolution) followed by a 1$\times$1-LCB, which enables feature extraction from different receptive fields. The second is an MS-Block for residual learning that further adjusts these features. 

\section{Experiment}
\subsection{Experimental Setup}
\subsubsection{Implementation Details}
For I-LIF neurons, the membrane time constant $\tau$ is initialized to 0.25. 
The models are trained on 4 NVIDIA A6000 GPUs, using the SGD optimizer with a learning rate of $0.01$. During post-processing, we use NMS, and the top 100 boxes are selected for evaluation.
Additionally, we use Complete IoU and Distribution Focal Loss for bounding box regression \cite{yolov8_ultralytics}, and Binary Cross Entropy for classification loss \cite{lin_focal_2017}.
For multiscale evaluation, following \cite{duan_centernet_2023}, we employ both original and horizontally flipped images at scale factors of 0.6, 1.0, 1.2, 1.5, and 1.8.

\subsubsection{Datasets}
COCO 2017 is a large-scale object detection benchmark with 118,287 training and 5,000 validation images \cite{lin_microsoft_2014}. Objects from 80 categories are annotated with classes and locations.
On this dataset, we train for 400 epochs with a batch size of 64 at a resolution of 640$\times$640, employing mosaic and mixup data augmentation \cite{chen_yoloms_2025}.

\subsubsection{Benchmark Settings}
We adopt Average Precision (AP) as the primary detection metric and firing rate to measure SNN neuronal activity. Formal definitions are provided in Appendix B-C. 
To evaluate computational cost, we use Floating-point Operations (FLOPs) for ANNs. Since SNN computation relies on sparse spike operations without multiplications, making FLOPs inapplicable, we employ Synaptic Operations (SOPs) instead. They are defined as:
\begin{equation}
\scalebox{0.90}{$\displaystyle
\text{FLOPs} = \sum_{n=1}^{N} 2 \times C^{n-1} \times C^{n} \times (K^{n})^{2} \times H^{n} \times W^{n},$}
\end{equation}
\begin{equation}\label{sop} 
\scalebox{0.87}{$\displaystyle
\text{SOPs} = \sum_{t=1}^{T} \sum_{n=1}^{N} {fr}^{t,n-1} \times C^{n-1} \times C^{n} \times (K^{n})^2 \times H^{n} \times W^{n},$}
\end{equation}
where $fr$ denotes the firing rate.
For the theoretical energy evaluation, we adopt the standard approach in the SNN field \cite{yao_scaling_2025, luo_integervalued_2025, yao_spikedriven_2024}, employing theoretical synaptic operation metrics independent of specific hardware platforms, to facilitate qualitative energy comparisons among different algorithms. Energy consumption is estimated as $E_{\text{SNNs}} = E_{\text{AC}} \times N_{\text{ACs}}$ and $E_{\text{ANNs}} = E_{\text{MAC}} \times N_{\text{MACs}}$, where $N_{\text{ACs}}$ and $N_{\text{MACs}}$ denote the number of accumulation and multiply-accumulate operations, respectively, with $E_{\text{AC}} = 0.9\,\text{pJ}$ and $E_{\text{MAC}} = 4.6\,\text{pJ}$. Since each SOP in SNNs involves only an accumulation, $\text{ACs} = \text{SOPs}$. For ANNs, $\text{FLOPs} = 2 \times \text{MACs}$.

\subsection{Local Firing Saturation Index}
\subsubsection{Definition}
To quantify the effectiveness of our method in addressing the local firing saturation problem, we define a Local Firing Saturation Index (LFSI). For the $n$-th layer with spike feature map $\mathbf{O} \in \mathbb{R}^{T \times C^n \times H^n \times W^n}$, a neuron at $(c,h,w)$ is \textbf{saturated} if it fires at maximum rate:
\begin{equation}\label{eq:saturation_def}
S_{\text{sat}}(\mathbf{O}^{c,h,w}) = 
\begin{cases}
1, & \text{if } \sum_{t=1}^{T} \mathbf{O}^{t,c,h,w} = T \times D, \\
0, & \text{otherwise}.
\end{cases}
\end{equation}
For each $S \times S$ spatial neighborhood $\mathcal{W}_{i,j}$ centered on location $(c,i,j)$, we compute local firing saturation density $\delta(c,i,j)$ by determining whether at least $\theta$ ($\theta \geq 2$) neurons within the neighborhood are simultaneously saturated. This directly reflects our definition of local firing saturation, where multiple spatially adjacent neurons reach saturation concurrently. The formulation is
\begin{equation}\label{eq_local_density}
\delta(c,i,j) = \mathbb{I}\left(\sum_{(p,q) \in \mathcal{W}_{i,j}} S_{\text{sat}}(\mathbf{O}^{c,p,q}) \geq \theta \right).
\end{equation}
The $\mathbb{I}(\cdot)$ is the indicator function that equals 1 when the condition holds and 0 otherwise.
The LFSI for a single layer is the average across all channels and spatial neighborhoods, 
\begin{equation}\label{eq_lsi_layer}
\text{LFSI}^{n} = \frac{1}{C^nH^nW^n} \sum_{c=1}^{C^n} \sum_{i=1}^{H^n} \sum_{j=1}^{W^n} \delta(c,i,j).
\end{equation}
For a network with $N$ spiking layers, the overall LFSI is
\begin{equation}\label{eq_lsi_overall}
\text{LFSI} = \frac{1}{N} \sum_{n=1}^{N} \text{LFSI}^n.
\end{equation} 
A higher LFSI indicates more severe local firing saturation, with $\text{LFSI} \in [0,1]$. Following the sensitivity analysis in Appendix B-D1, we adopt $S=3$ and $\theta=2$ as the default parameters, which ensure evaluation strictness while effectively capturing local firing saturation phenomena.

\subsubsection{Correlation Analysis of LFSI}
\begin{figure}[!t]
\centering
    \includegraphics[width=.95\columnwidth]{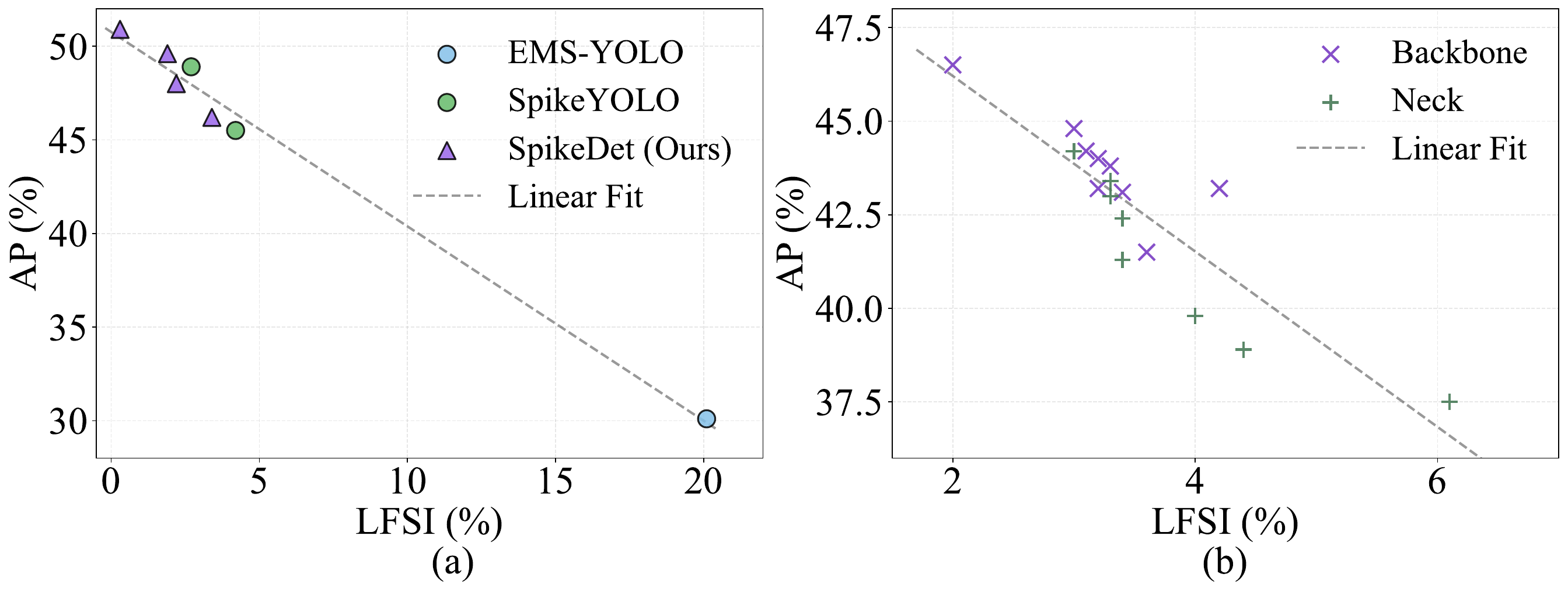}
\vspace{-0.2cm}
\caption{\textbf{Correlation analysis of LFSI and detection performance across different SNN-based object detectors.} (a) compares fully converged models, while (b) presents results from backbone and neck ablation studies.}
\label{fig_corr}
\vspace{-0.2cm}
\end{figure}

To validate the importance of addressing local firing saturation, we analyze the correlation between LFSI and detection performance across different model variants. As shown in Fig. \ref{fig_corr} (a), LFSI exhibits a strong negative correlation with AP (Pearson $r=-0.994$, $p<0.001$), demonstrating that mitigating local firing saturation is crucial for SNN-based object detection. 
Given the limited number of publicly available SNN-based detectors, we extend this analysis to our ablation study results in Fig. \ref{fig_corr} (b), where a consistent negative correlation is observed. Note that the ablation models adopt different training settings from the fully converged detectors and are therefore analyzed separately. Beyond performance correlation, Appendix B-D2 reveals a strong positive correlation between LFSI and firing rate, further demonstrating its validity as a metric.

\subsection{Ablation Studies}
We demonstrate the effectiveness of MDSNet and investigate how spiking neurons, MDS, and SMFM affect performance. All ablations use 150 epochs to ensure convergence for reliable comparison.  Ablations on downsampling in MDS, SF-Block, and time step appear in Appendix B-E.

\begin{table}[tbp]
\centering
\caption{\textbf{The ablation study on MDSNet.} 
The specific configuration of the proposed MDSNet is detailed in Appendix B-A.
}
\setlength{\tabcolsep}{3.4pt}
\begin{tabular}{lccccccc}
\hline
Backbone & Depth & AP & $\text{AP}_{50}$ & \makecell{Param \\ (M)} & \makecell{Firing \\ Rate(\%)} & \makecell{LFSI\\(\%)} & \makecell{Energy\\(mJ)} \\
\hline
EMS-ResNet \cite{su_deep_2023} & 34 & 43.2 & 59.5 & 22.9 & 14.6 & 4.2 & 22.7 \\
SEW-ResNet \cite{fang_deep_2021} & 34 & 43.2 & 59.7 & 22.9 & 12.2 & 3.2 & 51.1\\
MS-ResNet \cite{hu_advancing_2024} & 34 & 43.8 & 60.5 & 22.9 & 12.2 & 3.3 & 21.3\\
S-Backbone \cite{luo_integervalued_2025} & 47 & 44.0 & 60.6 & 24.2 & 12.2 & 3.2 & 30.4\\
\textbf{MDSNet} & \textbf{34} & \textbf{44.8} & \textbf{61.2} & \textbf{22.0} & \textbf{11.8} & \textbf{3.0} & \textbf{19.0}\\
\hline
MDSNet & 10 & 41.5 & 58.2 & 14.3 & 12.9 & 3.6 & 16.2\\
MDSNet & 18 & 43.1 & 59.5 & 17.5 & 12.4 & 3.4 & 17.5\\
\textbf{MDSNet} & \textbf{104} & \textbf{46.5} & \textbf{63.2} & \textbf{48.2} & \textbf{9.1} & \textbf{2.0} & \textbf{23.3}\\
\hline
\end{tabular}
\vspace{-0.2cm}
\label{tab_1}
\end{table}

\subsubsection{The Effectiveness of MDSNet}\label{exp_mds}
\begin{figure}[!t]
\centering
    \includegraphics[width=.9\columnwidth]{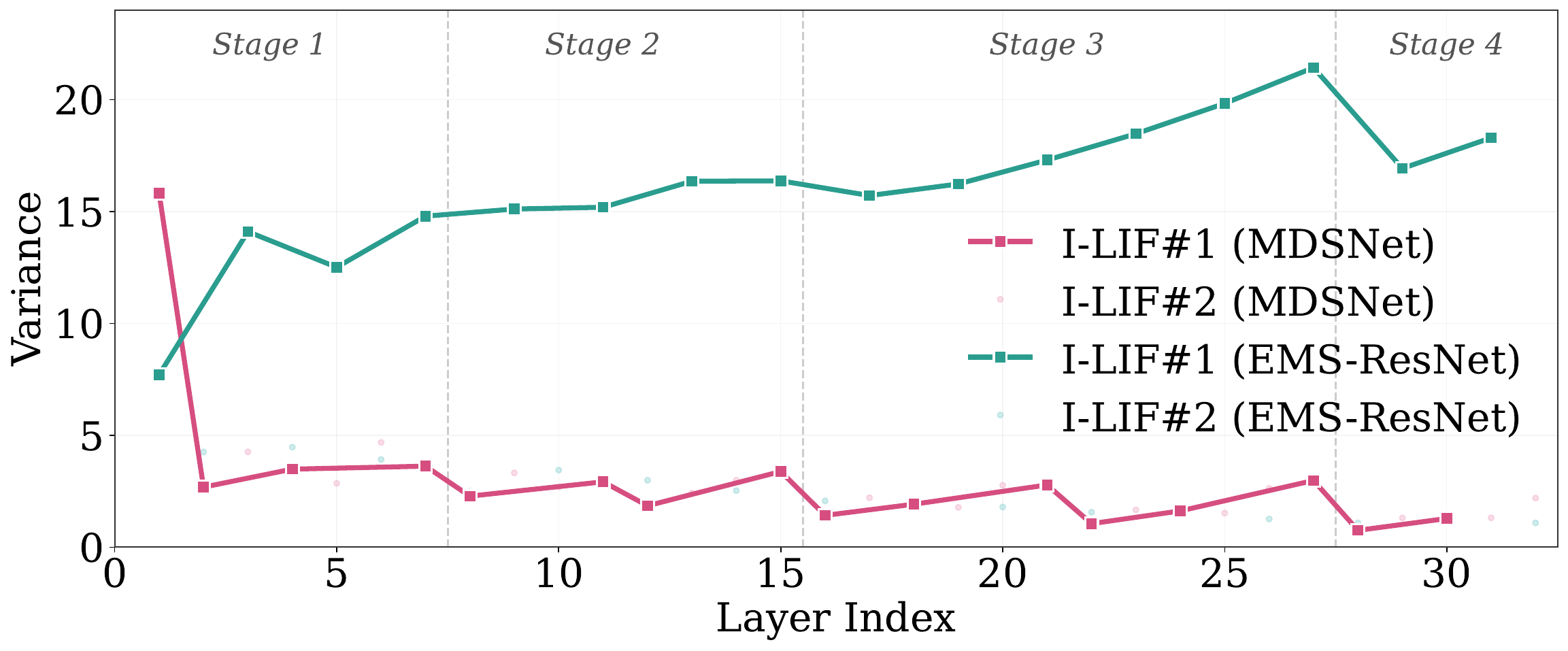}
\caption{\textbf{Presynaptic input variance across layers.}
MDSNet34 and EMS-ResNet34 are evaluated on COCO, where I-LIF\#1 and I-LIF\#2 denote the first and second residual-path neurons, respectively.
}
\label{fig_var}
\vspace{-0.2cm}
\end{figure}

To demonstrate MDSNet's feature extraction capabilities, we compare it with state-of-the-art SNN backbones.
For fair comparison, we pair each backbone with identical neck and detection head architectures.
Results are shown in rows 1 to 5 of Table \ref{tab_1}.
These results demonstrate that MDSNet34 significantly outperforms other models in AP and $\text{AP}_{50}$, while achieving lower LFSI, firing rate, and energy consumption.
Additionally, to verify the derivation in Section \ref{mdsnet_motivation}, we visualize the presynaptic input variance of EMS-ResNet and MDSNet in Fig. \ref{fig_var}. In contrast to EMS-ResNet, our MDSNet maintains stable presynaptic input variance throughout the network. For EMS-ResNet, the presynaptic input to I-LIF\#1 (i.e., the output of the previous residual block) exhibits gradually increasing variance, which is consistent with our theoretical analysis in Eq. \ref{var_sum}.

\subsubsection{Ablations on the Depth of MDSNet}
We experimentally validate that MDSNet scales to deeper architectures.
As shown in rows 5 to 8 of Table \ref{tab_1}, MDSNet's feature extraction capability improves with network depth. Notably, deeper networks exhibit lower LFSI and firing rates, suggesting that deeper architectures can effectively distribute neuronal activity across layers, mitigating local firing saturation.

\begin{table}[tp]
\centering
\caption{\textbf{Ablation study on MDS across models and spiking neurons.} \ding{55} in the MDS column denotes MDS replaced by membrane-based shortcuts. Ada. $V_{th}$ denotes a learnable adaptive threshold applied to spiking neurons.}
\setlength{\tabcolsep}{2.4pt}
\begin{tabular}{lcccccccc}
\hline
Model & MDS & T$\times$D & AP & $\text{AP}_{50}$ & \makecell{Param\\(M)} & \makecell{Firing \\ Rate(\%)} & \makecell{LFSI\\(\%)} & \makecell{Energy\\(mJ)} \\
\hline
\multirow{3}{*}{SpikeYOLO \cite{luo_integervalued_2025}} & \ding{55} & 4$\times$1 & 37.7 & 54.2 & 23.1 & 23.0 & 6.1 & 65.2 \\
    & \ding{55} & 1$\times$4 & 43.9 & 60.4 & 23.1 & 12.2 & 3.2 & 33.0 \\
    & \ding{51} & 1$\times$4 & 44.3 & 60.9 & 23.2 & 11.9 & 3.1 & 32.0 \\
\hline
\multirow{5}{*}{SpikeDet}  & \ding{51} & 4$\times$1 & 41.0 & 57.5 & 22.0 & 16.5 & 4.9 & 24.9 \\
    & \ding{55} & 4$\times$1 & 38.2 & 54.5 & 22.0 & 21.8 & 5.7 & 27.9 \\
    & \ding{51} & \textbf{1$\times$4} & \textbf{44.8} & \textbf{61.2} & \textbf{22.0} & \textbf{11.8} & \textbf{3.0} & \textbf{19.0} \\
    & \ding{55} & 1$\times$4 & 44.2 & 60.6 & 21.9 & 12.0 & 3.1 & 19.3 \\
w/ Ada. $V_{th}$ & \ding{51} & 1$\times$4 & 44.4 & 60.9 & 22.0 & 12.1 & 3.0 & 19.6 \\
\hline
\end{tabular}
\label{tab_2}
\vspace{-0.2cm}
\end{table}

\subsubsection{Ablations on MDS and Spiking Neurons}
To verify the contribution of MDS and the effect of spiking neurons on local firing saturation, we conduct ablation experiments as in Table \ref{tab_2}. 
First, under the default I-LIF (1$\times$4) setting, removing MDS degrades AP, LFSI, and energy consumption, while incorporating MDS into SpikeYOLO confirms its effectiveness and generalizability. 
Second, under the LIF (4$\times$1) setting, SpikeDet outperforms both SpikeDet without MDS and SpikeYOLO, demonstrating the significant structural contribution of MDS.
When I-LIF is introduced, SpikeYOLO achieves a notable AP improvement, partially compensating for the structural deficiency and narrowing the gap with SpikeDet. This indicates that I-LIF and MDS are complementary, the former indirectly alleviates local firing saturation by reducing quantization error, while the latter addresses it structurally.

Finally, we experiment with a learnable adaptive threshold on spiking neurons. Intuitively, it can reduce the probability of individual neurons reaching saturation, thereby alleviating local firing saturation. However, the results show a slight drop in AP with LFSI unchanged. This is because I-LIF adopts rectangular surrogate gradients, truncating the gradients of saturated neurons to zero. Consequently, the adaptive threshold receives no effective learning signal from the saturated regions requiring correction and thus fails to function as intended.

\begin{table*}[tbp]
\centering
\caption{\textbf{Performance Comparison with State-of-the-Art Models on COCO 2017 Validation Dataset.}
$\dagger$ indicates results from multiscale testing.
SNNs include FLOPs in the encoding layer. Therefore, in the computational cost column for SNNs, we report the results in the format of GFLOPs + GSOPs.
}
\begin{tabular}{llcccccccccccc}
\hline
Method & Model & AP & AP$_{50}$ & AP$_{75}$ & AP$_s$ & AP$_m$ & AP$_l$ & \makecell{Param\\(M)} & T$\times$D & \makecell{Firing\\Rate(\%)} & \makecell{LFSI\\(\%)} & \makecell{GFLOPs/ \\GSOPs} & \makecell{Energy\\(mJ)} \\
\hline
\multirow{13}{*}{ANNs} 
& RetinaNet$\dagger$ \cite{lin_focal_2017} & 41.8 & 62.9 & 45.7 & 25.6 & 45.1 & 54.1 & 56.7 & 1 & - & - & 330.2 & 759.5 \\
& TOOD \cite{feng_tood_2021} & 46.7 & 64.6 & 50.7 & 28.9 & 49.6 & 57.0 & 51.0 & 1 & - & - & 272.8 & 627.4 \\
& CenterNet$\dagger$ \cite{duan_centernet_2023} & 47.0 & 64.5 & 50.7 & 28.9 & 49.9 & 58.9 & 191.3 & 1 & - & - & 586.4 & 1348.8 \\
& UBT \cite{wang2025ubtransformer} & 51.9 & 69.6 & 56.3 & 35.4 & 55.1 & 65.3 & 50.8 & 1 & - & - & 293.0 & 673.9  \\
\cline{2-14}
\cline{2-14} 
& YOLOv8-N \cite{yolov8_ultralytics} & 37.2 & 52.7 & 40.3 & 18.9 & 40.5 & 52.5 & 3.2 & 1 & - & - & 8.7 & 20.2 \\
& YOLOv8-S \cite{yolov8_ultralytics} & 43.9 & 60.8 & 47.6 & 25.3 & 48.7 & 59.5 & 11.2 & 1 & - & - & 28.6 & 66.2 \\
& YOLOv8-M \cite{yolov8_ultralytics} & 49.8 & 66.9 & 54.2 & 32.6 & 54.9 & 65.9 & 25.9 & 1 & - & - & 78.9 & 182.2 \\
\cline{2-14}
\cline{2-14}
& YOLO-MS-XS \cite{chen_yoloms_2025} & 42.8 & 60.0 & 46.7 & 23.1 & 46.8 & 60.1 & 5.1 & 1 & - & - & 17.4 & 40.0 \\
& YOLO-MS-S \cite{chen_yoloms_2025} & 45.4 & 62.8 & 49.5 & 25.9 & 49.6 & 62.4 & 8.7 & 1 & - & - & 30.0 & 69.0 \\
& YOLO-MS \cite{chen_yoloms_2025} & 49.7 & 67.2 & 54.0 & 32.8 & 53.8 & 65.6 & 23.3 & 1 & - & - & 77.6 & 178.5 \\
\cline{2-14}
\cline{2-14}
& YOLOv12-N \cite{tian_yolov12_2025} & 40.6 & 56.7 & 43.8 & 20.2 & 45.2 & 58.2 & 2.6 & 1 & - & - & 6.5 & 15.0 \\
& YOLOv12-S \cite{tian_yolov12_2025} & 48.0 & 65.0 & 51.8 & 30.4 & 53.2 & 65.7 & 9.3 & 1 & - & - & 21.4 & 49.2 \\
& YOLOv12-M \cite{tian_yolov12_2025} & 52.5 & 69.6 & 57.1 & 35.9 & 58.2 & 68.8 & 20.2 & 1 & - & - & 67.5 & 155.3 \\
\hline
\multirow{17}{*}{SNNs} & Spiking-YOLO \cite{kim_spikingyolo_2020} & 25.7 & - & - & - & - & - & 10.2 & 3500$\times$1 & - & - & - & - \\
& EMS-YOLO \cite{su_deep_2023} & 30.1 & 50.1 & - & - & - & - & 33.9 & 4$\times$1 & 24.9 & 20.1 & 1.9 + 27.3 & 29.0 \\
& E-SpikeFormer \cite{yao_scaling_2025} & - & 58.8 & - & - & - & - & 38.7 & 1$\times$8 & - & - & 3.8 + 123.2 & 119.5 \\
& MHSANet-YOLO \cite{fan2025multisynaptic} & - & 66.9 & - & - & - & - & 76.3 & 1$\times$8 & - & - & - & 49.2 \\
\cline{2-14}
& SpikeYOLO-N \cite{luo_integervalued_2025} & 42.5 & 59.2 & - & - & - & - & 13.2 & 1$\times$4 & - & - & 0.4 + 24.7 & 23.1 \\
& SpikeYOLO-S \cite{luo_integervalued_2025} & 45.5 & 62.3 & 49.2 & 25.5 & 50.4 & 61.3 & 23.1 & 1$\times$4 & 12.8 & 4.2 & 0.5 + 37.2 & 34.6 \\
& SpikeYOLO-M \cite{luo_integervalued_2025} & 47.4 & 64.6 & - & - & - & - & 48.1 & 1$\times$4 & - & - & 0.6 + 74.6 & 68.5 \\
& SpikeYOLO-L \cite{luo_integervalued_2025} & 48.9 & 66.2 & 53.4 & 29.3 & 55.0 & 64.2 & 68.8 & 1$\times$4 & 10.4 & 2.7 & 0.7 + 91.7 & 84.2 \\
& SpikeYOLO-L$\dagger$ \cite{luo_integervalued_2025} & 51.2 & 68.0 & 56.7 & 36.4 & 56.7 & 65.6 & 68.8 & 1$\times$4 & 10.4 & 2.7 & 0.7 + 91.7 & 84.2 \\
\cline{2-14}
& SpikeDet-S & 46.2 & 62.6 & 50.5 & 26.8 & 51.2 & 60.8 & 22.0 & 1$\times$4 & 12.3 & 3.4 & 1.0 + 19.3 & 19.6 \\
& SpikeDet-S$\dagger$ & 48.2 & 65.3 & 53.6 & 32.4 & 53.2 & 62.2 & 22.0 & 1$\times$4 & 12.3 & 3.4 & 1.0 + 19.3 & 19.6 \\
& SpikeDet-M & 48.0 & 64.8 & 52.1 & 27.5 & 54.1 & 63.6 & 48.2 & 1$\times$4 & 9.6 & 2.2 & 1.0 + 24.8 & 24.6 \\
& SpikeDet-M$\dagger$ & 50.2 & 67.5 & 55.5 & 34.1 & 55.9 & 64.3 & 48.2 & 1$\times$4 & 9.6 & 2.2 & 1.0 + 24.8 & 24.6 \\
& SpikeDet-L & 49.6 & 66.5 & 54.3 & 29.6 & 55.5 & 65.0 & 75.2 & 1$\times$4 & 8.8 & 1.9 & 1.2 + 34.0 & 33.4 \\
& SpikeDet-L$\dagger$ & 51.4 & 68.6 & 56.9 & 36.6 & 56.8 & 65.4 & 75.2 & 1$\times$4 & 8.8 & 1.9 & 1.2 + 34.0 & 33.4 \\
& \textbf{SpikeDet-X} & \textbf{50.9} & \textbf{67.6} & \textbf{54.7} & \textbf{31.3} & \textbf{57.1} & \textbf{66.5} & \textbf{75.2} & \textbf{1$\times$8} & \textbf{4.3} & \textbf{0.3} & \textbf{1.2 + 41.7} & \textbf{40.4} \\  
& \textbf{SpikeDet-X$\dagger$} & \textbf{52.2} & \textbf{69.3} & \textbf{57.2} & \textbf{36.9} & \textbf{57.2} & \textbf{65.9} & \textbf{75.2} & \textbf{1$\times$8} & \textbf{4.3} & \textbf{0.3} & \textbf{1.2 + 41.7} & \textbf{40.4} \\ 
\hline
\end{tabular}
\label{tab_5}
\vspace{-0.2cm}
\end{table*}

\begin{table}[tbp]
\centering
\caption{\textbf{Ablations on Number of Fusion Directions in SMFM.}}
\setlength{\tabcolsep}{3.1pt}
\begin{tabular}{lccccccc}
\hline
\makecell[l]{Fusion \\ Method} & \makecell{Fusion \\ Directions} & AP & $\text{AP}_{50}$ & \makecell{Param \\ (M)} & \makecell{Firing \\ Rate(\%)} & \makecell{LFSI\\(\%)} & \makecell{Energy\\(mJ)} \\
\hline
SMFM & 1 & 41.3 & 58.6 & 21.1 & 13.1 & 3.4 & 19.5 \\
SMFM & 2 & 43.0 & 59.9 & 21.6 & 12.2 & 3.3 & 17.6 \\
\textbf{SMFM} & \textbf{4} & \textbf{44.8} & \textbf{61.2} & \textbf{22.0} & \textbf{11.8} & \textbf{3.0} & \textbf{19.0} \\
SMFM & 6 & 44.2 & 60.7 & 22.1 & 11.3 & 3.0 & 19.6 \\
\hline
FPN \cite{lin_feature_2017} & 1 & 38.9 & 56.4 & 14.9 & 14.2 & 4.4 & 13.2 \\
PAN \cite{liu_path_2018} & 2 & 39.8 & 57.7 & 17.2 & 12.9 & 4.0 & 12.3 \\
BiFPN-D0 & 6 & 37.5 & 55.0 & 11.2 & 18.2 & 6.1 & 8.6 \\
BiFPN-D6 \cite{tan_efficientdet_2020} & 16 & 42.4 & 59.9 & 25.0 & 11.7 & 3.4 & 18.9 \\
S-PAN \cite{luo_integervalued_2025} & 2 & 43.4 & 60.3 & 21.0 & 12.0 & 3.3 & 26.4 \\
\hline
\end{tabular}
\label{tab_3}
\vspace{-0.2cm}
\end{table}

\subsubsection{Ablations on Number of Fusion Directions in SMFM}
Rows 1 to 4 of Table \ref{tab_3} compare the impact of different fusion direction numbers in SMFM. For fair comparison, network width is adjusted to maintain comparable parameters across models. Both LFSI and firing rate consistently decrease as fusion directions increase, validating our analysis in Section \ref{SMFM} that more fusion directions help mitigate local firing saturation. 
However, detection performance begins to degrade beyond 4 directions, indicating that excessive structural complexity hinders optimization. Therefore, we adopt 4 fusion directions for SMFM.

We further investigate classical feature fusion modules from both ANN and SNN detectors \cite{lin_feature_2017, liu_path_2018, tan_efficientdet_2020, luo_integervalued_2025}. We adapt ANN methods for SNNs by replacing activations with I-LIF neurons. As shown in rows 5 to 9 of Table \ref{tab_3}, similar trends are observed, more fusion directions consistently yield better suppression of local firing saturation. 

\subsection{Comparisons with State-of-The-Art Models}\label{comp_coco}
\begin{figure}[t]
\centering
\includegraphics[width=1.\columnwidth]{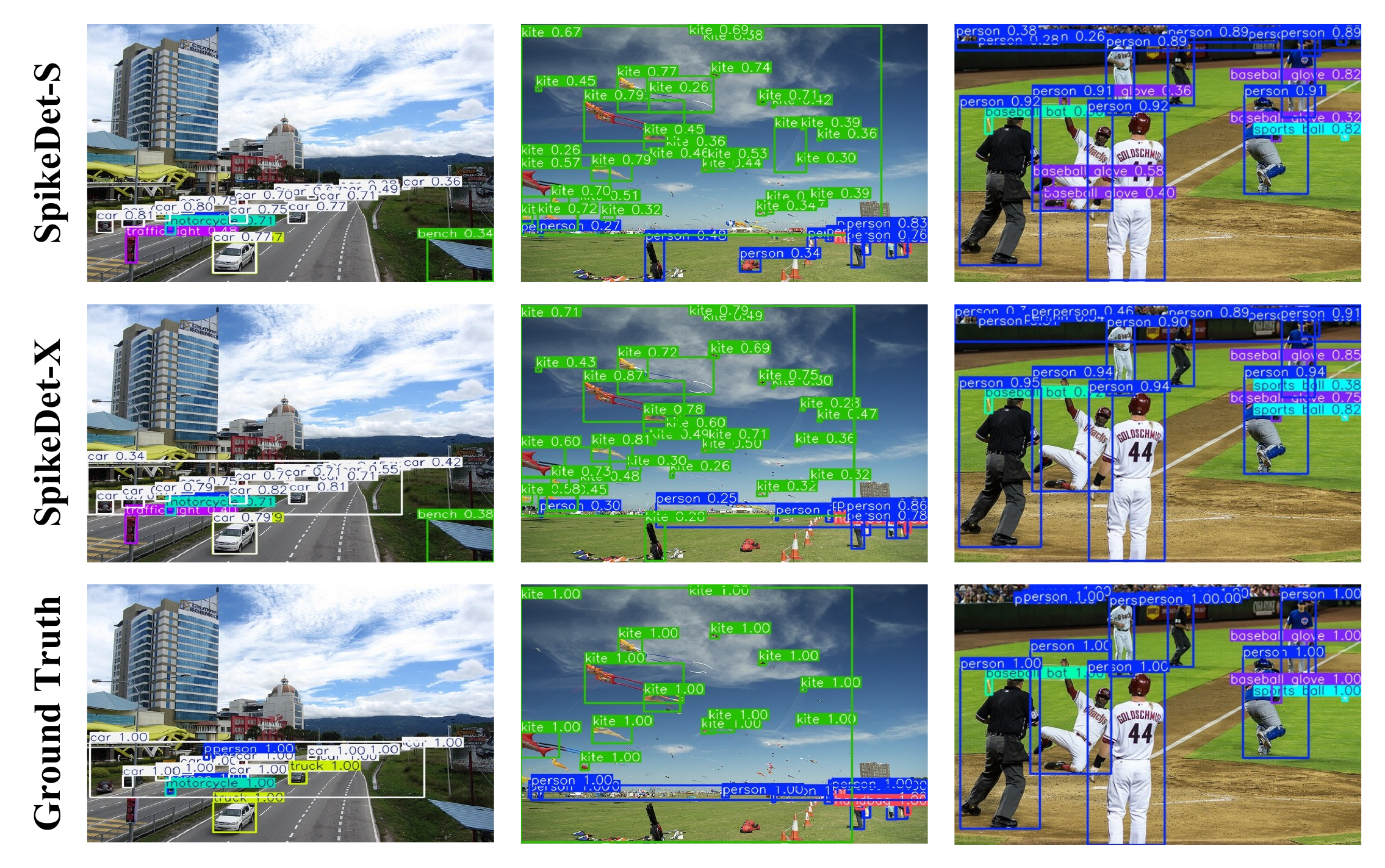}
\caption{\textbf{Detection results on the COCO 2017 dataset.}}
\label{fig_7}
\vspace{-0.25cm}
\end{figure}

Based on ablation studies, we propose SpikeDet-S and SpikeDet-M using MDSNet34 and MDSNet104 as backbones, respectively. Furthermore, to explore the performance upper limit of SNN-based detectors, we widen SpikeDet-M to propose SpikeDet-L. We further increase T$\times$D from 1$\times$4 to 1$\times$8, resulting in SpikeDet-X. The comparison results with state-of-the-art methods are shown in Table \ref{tab_5}.
Compared to other SNN-based object detectors, our model achieves \textbf{52.2\%} AP and \textbf{69.3\%} $\text{AP}_{50}$, representing \textbf{+3.3\%} and \textbf{+3.1\%} improvements over the previous best (48.9\% AP, 66.2\% $\text{AP}_{50}$). Remarkably, this is achieved with only $\mathbf{\frac{1}{2}}$ the energy consumption, further strengthening the energy efficiency advantage of SNNs.
Moreover, our method reduces LFSI by approximately 30\% compared to SpikeYOLO at similar parameter scales, which further validates our analysis that local firing saturation constrains the advancement of SNNs in performance and energy consumption.
Furthermore, compared to ANN-based methods, SpikeDet not only further narrows the gap between SNNs and ANNs, but also demonstrates a better trade-off between accuracy and energy consumption. For example, SpikeDet-X achieves comparable accuracy to YOLOv12-M while providing \textbf{3.8}$\times$ the energy efficiency. Inference results at various scales are shown in Fig. \ref{fig_7}.

\subsection{Evaluation on Object Detection Sub-Tasks}
We further apply SpikeDet to four object detection sub-tasks: event camera on the GEN1 dataset \cite{tournemire_large_2020}, underwater on the URPC 2019 dataset \cite{wang_ulo_2022}, low-light on the ExDARK dataset \cite{loh2019getting}, and dense scene on the CrowdHuman dataset \cite{shao2018crowdhuman}.
These scenarios evaluate the generalization and robustness of SpikeDet across diverse conditions. Implementation details and dataset descriptions are in Appendix B-B.

\begin{table}[tbp]
\centering
\caption{\textbf{Performance Comparison with State-of-the-Art Models on GEN1 Dataset.}}
\setlength{\tabcolsep}{3.0pt}  
\begin{tabular}{lcccccccc}
\hline
Model & AP & AP$_{50}$ & \makecell{Param\\(M)} & T$\times$D & \makecell{Firing\\Rate(\%)} & \makecell{LFSI\\(\%)} & \makecell{Energy\\(mJ)} \\
\hline
RVT \cite{gehrig_recurrent_2023} & 47.2 & - & 18.5 & 10 & - & - & 116.7 \\
S5-Vit-B \cite{zubic_state_2024} & 47.4 & - & 18.2 & 10 & - & - & - \\
\hline
EMS-YOLO \cite{su_deep_2023} & 31.0 & 59.0 & 14.4 & 5$\times$1 & 17.8 & - & 3.4 \\
SFOD \cite{fan_sfod_2024} & 32.1 & - & 11.9 & 5$\times$1 & 24.4 & 13.8  & 7.3 \\
EAS-SNN \cite{wang_eassnn_2025} & 37.5 & 69.9 & 25.3 & 5$\times$1 & 20.3 & - & 28.1 \\
SpikeYOLO-S \cite{luo_integervalued_2025} & 38.9 & 66.4 & 23.1 & 5$\times$1 & 10.5 & 4.6 & 19.7 \\
SpikeYOLO-S \cite{luo_integervalued_2025} & 40.4 & 67.2 & 23.1 & 4$\times$2 & 14.0 & 5.4 & 12.9 \\
\textbf{SpikeDet-S} & \textbf{46.5} & \textbf{69.2} & \textbf{22.0} & \textbf{5$\times$1} & \textbf{10.3} & \textbf{2.0} & \textbf{10.7} \\
\textbf{SpikeDet-S} & \textbf{47.6} & \textbf{70.1} & \textbf{22.0} & \textbf{4$\times$2} & \textbf{13.7} & \textbf{4.2} & \textbf{9.9} \\
\hline
\end{tabular}
\label{tab_6}
\vspace{-0.2cm}
\end{table}

\subsubsection{Event Camera Object Detection}
As shown in Table \ref{tab_6}, our method significantly outperforms both SNN-based and ANN-based object detectors on the GEN1 dataset. Importantly, this is achieved while maintaining the energy efficiency advantage of SNNs.
This improvement stems from the unique characteristics of event-based data. Inherently sparse event streams are already challenging to recognize, and local firing saturation further exacerbates this difficulty. Therefore, addressing local firing saturation is particularly crucial for event-based object detection. SpikeDet achieves LFSI values approximately 1.3$\times$ and 3.3$\times$ lower than those of SpikeYOLO and SFOD, respectively, demonstrating the effectiveness of our approach.
Notably, when adjusting T$\times$D from 5$\times$1 to 4$\times$2, performance improves despite an increase in LFSI. This does not contradict our earlier analysis, as Fig.~\ref{fig_corr} compares models under the same neuron type and similar T$\times$D products.

\subsubsection{Underwater, Low-Light, and Dense Scene Object Detection}
As shown in Table \ref{tab_7}, SpikeDet consistently achieves strong performance across all three datasets.
On URPC 2019, SpikeDet outperforms YOLOv9-S-UI by 1.4\% AP while reducing energy consumption by 5.7 mJ. The substantial LFSI reduction enhances feature discrimination for low-contrast and blurred underwater objects.
On ExDARK, SpikeDet achieves a 2.6\% AP improvement and 18.9 mJ energy reduction compared to SpikeYOLO, demonstrating robust performance under low-light conditions and a better accuracy-energy trade-off than ANN-based methods.
On CrowdHuman, SpikeDet outperforms SpikeYOLO by 4.3\% AP with 19.2 mJ less energy, where lower LFSI enables better discrimination of overlapping objects in crowded scenes.

\begin{table}[!t]
\centering
\caption{\textbf{Performance Comparison with State-of-the-Art Models on URPC 2019, ExDARK, and CrowdHuman Datasets.} $\dagger$ indicates results from multiscale testing.}
\setlength{\tabcolsep}{2.5pt}
\begin{tabular}{clccccccc}
\hline
& Model & AP & AP$_{50}$ & \makecell{Param\\(M)} & T$\times$D & \makecell{Firing\\Rate(\%)} & \makecell{LFSI\\(\%)} & \makecell{Energy\\(mJ)} \\
\hline
\multirow{8}{*}{\rotatebox{90}{\textit{URPC 2019}}}
 & YOLOv8-N \cite{yolov8_ultralytics}            & 43.6 & 77.7 & 3.2  & 1          & -    & -   & 5.0  \\
 & YOLOv9-S-UI \cite{pan_optimization_2024}      & 48.7 & 78.1 & 4.1  & 1          & -    & -   & 10.7 \\
\cline{2-9}
 & Spiking-YOLO \cite{kim_spikingyolo_2020}      & 22.4 & 53.7 & 8.7  & 3500       & -    & -   & -    \\
 & SU-YOLO \cite{li_suyolo_2025}                 & 42.9 & 78.8 & 7.0  & 4          & -    & -   & 3.0  \\
 & SpikeYOLO-S \cite{luo_integervalued_2025}     & 47.9 & 83.8 & 23.1 & 1$\times$4 & 12.2 & 2.4 & 8.6  \\
 & \textbf{SpikeDet-S}                           & \textbf{49.2} & \textbf{84.3} & \textbf{22.0} & \textbf{1$\times$4} & \textbf{9.9}  & \textbf{1.3} & \textbf{5.0} \\
 & \textbf{SpikeDet-S$\dagger$}                  & \textbf{50.1} & \textbf{86.1} & \textbf{22.0} & \textbf{1$\times$4} & \textbf{9.9}  & \textbf{1.3} & \textbf{5.0} \\
\hline
\multirow{5}{*}{\rotatebox{90}{\textit{ExDARK}}}
 & DETR-R50 \cite{carion_endtoend_2020}          & 25.0 & 55.0 & 41.3 & 1          & -    & -   & 219.0 \\
 & YOLOv8-S \cite{yolov8_ultralytics}            & 40.7 & 66.2 & 11.1 & 1          & -    & -   & 65.6  \\
\cline{2-9}
 & SpikeYOLO-S \cite{luo_integervalued_2025}     & 40.4 & 65.1 & 23.1 & 1$\times$4 & 12.0 & 2.5 & 36.1  \\
 & \textbf{SpikeDet-S}                           & \textbf{42.1} & \textbf{66.5} & \textbf{22.0} & \textbf{1$\times$4} & \textbf{10.1} & \textbf{1.1} & \textbf{17.2} \\
 & \textbf{SpikeDet-S$\dagger$}                  & \textbf{43.0} & \textbf{69.6} & \textbf{22.0} & \textbf{1$\times$4} & \textbf{10.1} & \textbf{1.1} & \textbf{17.2} \\
\hline
\multirow{5}{*}{\rotatebox{90}{\textit{CrowdHuman}}}
 & DETR-R50 \cite{carion_endtoend_2020}          & 27.8 & 55.5 & 41.3 & 1          & -    & -   & 219.0 \\
 & YOLOv8-S \cite{yolov8_ultralytics}            & 51.7 & 79.3 & 11.1 & 1          & -    & -   & 65.3  \\
\cline{2-9}
 & SpikeYOLO-S \cite{luo_integervalued_2025}     & 50.4 & 78.7 & 23.1 & 1$\times$4 & 11.2 & 2.1 & 37.6  \\
 & \textbf{SpikeDet-S}                           & \textbf{52.8} & \textbf{78.8} & \textbf{22.0} & \textbf{1$\times$4} & \textbf{10.1} & \textbf{1.4} & \textbf{18.4} \\
 & \textbf{SpikeDet-S$\dagger$}                  & \textbf{54.7} & \textbf{80.4} & \textbf{22.0} & \textbf{1$\times$4} & \textbf{10.1} & \textbf{1.4} & \textbf{18.4} \\
\hline
\end{tabular}
\label{tab_7}
\end{table}

\begin{table}[!t]
\centering
\caption{\textbf{Hardware Deployment Comparison.}}
\setlength{\tabcolsep}{3.3pt}
\begin{tabular}{lcccccc}
\hline
Hardware & Model & AP & FPS & \makecell{Power\\(W)} & FPS/W \\
\hline
\multirow{2}{*}{NVIDIA Xavier SoC} 
  & YOLOv8-S & 43.9 & 41.3 & 9.8 & 4.2 \\
  & YOLOv8-M & 49.8 & 21.0 & 11.5 & 1.8 \\
\hline
\multirow{2}{*}{\textbf{Lynxi KA200}} 
  & \textbf{SpikeDet-S} & \textbf{46.2} & \textbf{68.2} & \textbf{7.1} & \textbf{9.6} \\
  & \textbf{SpikeDet-M} & \textbf{48.0} & \textbf{41.6} & \textbf{8.3} & \textbf{5.0} \\
\hline
\end{tabular} 
\label{tab_hardware}
\vspace{-0.2cm}
\end{table}

\subsection{Hardware Deployment}
Since theoretical energy evaluation overlooks hardware-related overhead such as memory access, we further conduct real hardware deployment. 
As SNNs are inherently optimized for neuromorphic hardware while ANNs are designed for GPU-based platforms, deploying each model on its native hardware ensures a fair comparison reflecting each paradigm's true efficiency.
For SNN deployment, we use the DF400 edge device equipped with the Lynxi KA200 neuromorphic chip, designed to exploit the sparse spike-driven computation of SNNs. For ANN deployment, we use the NVIDIA Jetson Xavier NX (16 GB), the most typical GPU-based edge platform for ANN inference. We deploy SpikeDet and YOLOv8 of S and M scales on the two platforms, respectively. Implementation details are in Appendix B-A. As shown in Table \ref{tab_hardware}, SpikeDet-S and M achieve 2.3 $\times$
and 2.8 $\times$ higher FPS/W than their YOLOv8 counterparts, respectively. This aligns with our theoretical energy analysis, as larger SNN models exhibit lower firing rates and LFSI, enabling sparser computation and greater energy efficiency advantages.

\section{Conclusion}
In this paper, we identify the local firing saturation problem in existing SNN-based object detectors, which limits both accuracy and energy efficiency. To address this, we propose SpikeDet, a novel SNN-based object detector.
The key contribution is MDSNet, which effectively stabilizes synaptic input distribution and mitigates local firing saturation through the proposed MDS and architectural improvements. For feature fusion, we propose SMFM, which employs a multi-directional fusion strategy to enhance multi-scale detection while preserving neuron firing patterns. We also introduce the Local Firing Saturation Index (LFSI) to quantify the effectiveness of our approach in addressing local firing saturation.
Our model outperforms other SNN-based models in accuracy on COCO 2017 and generalizes well to diverse downstream tasks. Meanwhile, it maintains significantly lower energy consumption, achieving an optimal balance between accuracy and energy efficiency.

\appendices
\section{Proof of the Propositions}\label{proof}
\subsection{Proof of Proposition 1}\label{proof2}
\noindent\textbf{Proposition 1.}
\textit{For $d$ stacked LCB layers ($d \geq 1$) with arbitrary kernel sizes, under the convolution zero-mean weight and zero bias initialization, the output at the $(n+d)$-th layer $\mathbf{y}^{t,n+d}$ is uncorrelated with the input at the $n$-th layer $\mathbf{x}^{t,n}$.}

\noindent\rule{\columnwidth}{0.5pt} 

We first analyze the case of a single LCB, conducting our analysis according to the processing sequence of its internal components. The first component is the I-LIF neuron. It receives input $\mathbf{x}^{t,n}$ and produces output spikes $\mathbf{o}^{t,n} = \mathit{Clip} \left( \mathit{round} (\mathbf{u}^{t,n}), 0, D \right)$.

Subsequently, we combine the convolution with tdBN following \cite{zheng_going_2021}. During training, tdBN normalizes the pre-synaptic inputs along the $n$-th channel dimension as,
\begin{equation}
\hat{\mathbf{x}}_n = \frac{\alpha V_{th}(\mathbf{x}_n - E[\mathbf{x}_n])}{\sqrt{\text{Var}[\mathbf{x}_n] + \epsilon}},
\end{equation}
\begin{equation}
\mathbf{y}_n = \lambda_n \hat{\mathbf{x}}_n + \beta_n,
\end{equation}
where $E[\mathbf{x}_n]$ and $\text{Var}[\mathbf{x}_n]$ are estimated over the mini-batch across all time steps, $\lambda_n$ and $\beta_n$ are learnable parameters, $\alpha$ is a hyperparameter. During inference, tdBN is merged into the convolution to maintain a full-spiking network, where the merged weight and bias are:
\begin{equation}
W^{\prime}_{m,n} = \lambda_n \frac{\alpha V_{th} W_{m,n}}{\sqrt{\sigma^2_{inf,n} + \epsilon}},
\end{equation}
\begin{equation}\label{merged_bias}
B^{\prime}_{m,n} = \lambda_n \frac{\alpha V_{th}(B_{m,n} - \mu_{inf,n})}{\sqrt{\sigma^2_{inf,n} + \epsilon}} + \beta_n.
\end{equation}
Here $W_{m,n}$ and $B_{m,n}$ denote the weight and bias mapping from the $m$-th feature map of the previous layer to the $n$-th feature map of the next layer, and the superscript $\prime$ indicates the merged version. $\sigma^2_{inf,n}$ and $\mu_{inf,n}$ are the dataset-level estimates of $\text{Var}[\mathbf{x}_n]$ and $E[\mathbf{x}_n]$, respectively.

Taking the expectation of the merged weight:
\begin{equation}\label{weight_mean}
E[W^{\prime}_{m,n}] = \lambda_n \frac{\alpha V_{th}}{\sqrt{\sigma^2_{inf,n} + \epsilon}} E[W_{m,n}].
\end{equation}
Since $E[W_{m,n}] = 0$ (zero-mean weight initialization) and $\lambda_n, \alpha, V_{th}, \sigma^2_{inf,n}, \epsilon$ are deterministic constants:
\begin{equation}\label{weight_mean_zero}
E[W^{\prime}_{m,n}] = 0.
\end{equation}
Thus, tdBN merging preserves the zero-mean property of weights.
Furthermore, given that $B_{m,n}$ is initialized to zero, and according to the default parameters in \cite{zheng_going_2021}, $\mu_{inf,n}$ and $\beta_n$ are also 0, so from Eq. \ref{merged_bias}, we get $B^{\prime}_{m,n} = \mathbf{0}$.

As in \cite{he2015delving}, we assume that the elements in $\mathbf{x}^{t,n}$ and $W_{m,n}$ are independent and identically distributed (i.i.d.). For rigorous covariance analysis, we consider individual tensor elements. Let $y_i^{t,n}$, $x_i^{t,n}$, $w_i^{t,n}$, $b_i^{t,n}$, and $o_i^{t,n}$ denote arbitrary elements of $\mathbf{y}^{t,n}$, $\mathbf{x}^{t,n}$, $W^{\prime}_{m,n}$, $B^{\prime}_{m,n}$, and $\mathbf{o}^{t,n}$, respectively. 
The $\mathbf{y}^{t,n}$ represents the output of a single LCB, and its element $y_i^{t,n}$ can be expressed as
\begin{equation}\label{conv_element}
y_i^{t,n} = \sum_{j \in \text{RF}(i)} w_j^n \cdot o_j^n + b_i^{t,n},
\end{equation}
where $\text{RF}(i)$ denotes the receptive field corresponding to output position $i$. For simplicity,
\begin{equation}\label{conv_simple1}
y_i^{t,n} = \sum_j w_j^{t,n} \cdot o_j^{t,n} + b_i^{t,n}.
\end{equation}
From the above derivation, we can obtain $b_i^{t,n} = 0$, therefore,
\begin{equation}\label{conv_simple2}
y_i^{t,n} = \sum_j w_j^{t,n} \cdot o_j^{t,n}.
\end{equation}

Then, we derive the covariance between the output $y_i^{t,n}$ and the input $x_i^{t,n}$,
\begin{equation}\label{cov_def}
\text{Cov}(y_i^{t,n}, x_i^{t,n}) = E[y_i^{t,n} \cdot x_i^{t,n}] - E[y_i^{t,n}] E[x_i^{t,n}].
\end{equation}
Since $E[x_i^{t,n}] = 0$,
\begin{equation}\label{cov_simplified}
\text{Cov}(y_i^{t,n}, x_i^{t,n}) = E[y_i^{t,n} \cdot x_i^{t,n}].
\end{equation}
Expanding the above equation, we get
\begin{equation}\label{expand_E}
E[y_i^{t,n} \cdot x_i^{t,n}] = E\left[\left(\sum_j w_j^{t,n} \cdot o_j^{t,n}\right) \cdot x_i^{t,n}\right].
\end{equation}
Exchanging summation and expectation,
\begin{equation}\label{expand_sum}
E[y_i^{t,n} \cdot x_i^{t,n}] = \sum_j E[w_j^{t,n} \cdot o_j^{t,n} \cdot x_i^{t,n}].
\end{equation}
Since weights $w_j^{t,n}$ are independent of inputs at initialization,
\begin{equation}\label{independence}
E[w_j^{t,n} \cdot o_j^{t,n} \cdot x_i^{t,n}] = E[w_j^{t,n}] \cdot E[o_j^{t,n} \cdot x_i^{t,n}].
\end{equation}
Therefore:
\begin{equation}\label{term_zero}
E[w_j^{t,n} \cdot o_j^{t,n} \cdot x_i^{t,n}] = 0 \cdot E[o_j^{t,n} \cdot x_i^{t,n}] = 0.
\end{equation}
This holds for all $j$.
From Eq. \ref{expand_sum}:
\begin{equation}\label{E_final}
E[y_i^{t,n} \cdot x_i^{t,n}] = \sum_j E[w_j^{t,n} \cdot o_j^{t,n} \cdot x_i^{t,n}] = \sum_j 0 = 0.
\end{equation}
\begin{equation}\label{cov_zero}
\text{Cov}(y_i^{t,n}, x_i^{t,n}) = 0.
\end{equation}
Therefore, for a single LCB, its output $\mathbf{y}$ is uncorrelated with the input $\mathbf{x}$.

Finally, we consider $d$ stacked LCBs ($d \geq 1$) and analyze them using mathematical induction. Assume that for $(d-1)$ stacked LCBs, we have $\text{Cov}(y^{t,n+d-1}, x^{t,n}) = 0$. We prove that this also holds for $d$ stacked LCBs, i.e., $\text{Cov}(y^{t,n+d}, x^{t,n}) = 0$.

The output of layer $(n+d)$ is:
\begin{equation}
y^{t,n+d}_i = \sum_j w^{t,n+d}_j \cdot o^{t,n+d}_j.
\end{equation}
Since $E[\mathbf{x}^{t,n}] = 0$,
\begin{equation}
\begin{aligned}
\text{Cov}(y^{t,n+d}_i, x^{t,n}_i) &= E[y^{t,n+d}_i \cdot x^{t,n}_i]\\
&= \sum_j E[w^{t,n+d}_j \cdot o^{t,n+d}_j \cdot x^{t,n}_i].\\
\end{aligned}
\end{equation}
At initialization, weights $w^{t,n+d}_j$ are independent of the input $\mathbf{x}^{t,n}$ at layer $n$. Although $o^{t,n+d}_j$ depends on $\mathbf{x}^{t,n}$ through intermediate layers, the weight independence holds:
\begin{equation}
E[w^{t,n+d}_j \cdot o^{t,n+d}_j \cdot x^{t,n}_i] = E[w^{t,n+d}_j] \cdot E[o^{t,n+d}_j \cdot x^{t,n}_i].
\end{equation}
Since $E[w^{t,n+d}_j] = 0$ by zero-mean initialization,
\begin{equation}
E[w^{t,n+d}_j \cdot o^{t,n+d}_j \cdot x^{t,n}_i] = 0.
\end{equation}
Thus $\text{Cov}(y^{t,n+d}_i, x^{t,n}_i) = 0$. By mathematical induction, $\text{Cov}(\mathbf{y}^{t,n+d}, \mathbf{x}^{t,n}) = 0$ for any starting layer $n$ and $d$ stacked LCB layers. This completes the proof of Proposition 1.

\subsection{Proof of Proposition 2}\label{proof3}
\noindent\textbf{Proposition 2.}
\textit{For I-LIF neurons, the variance of the membrane potential is proportional to the variance of the input, i.e., $\text{Var}[\mathbf{u}^{t,n}] \propto \text{Var}[\mathbf{x}^{t,n}]$.}

\noindent\rule{\columnwidth}{0.5pt}
  
From Eqs. 1 and 3 in the main paper, we obtain:
\begin{equation}\label{proof3_1}
\begin{aligned}
\mathbf{u}^{t,n} &= \tau (\mathbf{u}^{t-1,n} - \mathbf{o}^{t-1,n}) + \mathbf{x}^{t,n-1}\\
&= \tau [\tau(\mathbf{u}^{t-2,n} - \mathbf{o}^{t-2,n}) + \mathbf{x}^{t-1,n-1} - \mathbf{o}^{t-1,n}] + \mathbf{x}^{t,n-1}\\
&= {\tau}^{t}\mathbf{u}^{0,n} - \sum_{p=1}^{t-1} {\tau}^{t-p}\mathbf{o}^{p,n} + \sum_{p=1}^{t} {\tau}^{t-p}\mathbf{x}^{p,n-1}.\\
\end{aligned}
\end{equation}
Since I-LIF can fire integer-valued spikes, each spike emission reduces the membrane potential to a level close to 0. Therefore, we can assume that the last spike is fired at time step $t^{\prime}$, and thus ignore the influence of $\mathbf{o}$ on the membrane potential, yielding:
\begin{equation}\label{proof3_2}
\mathbf{u}^{t,n} = {\tau}^{t}\mathbf{u}^{0,n} + \sum_{p=t^{\prime}}^{t} {\tau}^{t-p}\mathbf{x}^{p,n-1}.
\end{equation}

Moreover, since $\tau$ is a tiny constant, which is only 0.25 in our implementation, the above equation can be simplified to:
\begin{equation}\label{proof3_3}
\mathbf{u}^{t,n} = \mathbf{x}^{t,n-1} + {\tau}\mathbf{x}^{t-1,n-1}.
\end{equation}
Assume that $\mathbf{x}$ is independently and identically distributed for each time step $t$. Thus, we can conclude that the membrane potential variance is proportional to the input variance, i.e., $\text{Var}[\mathbf{u}^{t,n}] \propto \text{Var}[\mathbf{x}^{t,n-1}]$.

\begin{table}[t]
\centering
\caption{\textbf{The Structure of MDSNet for COCO 2017.} 
The numbers in parentheses indicate the number of MS-Blocks contained in the residual path of the MDS-Block.}
\resizebox{\columnwidth}{!}{%
\begin{tabular}{c|c|c|c|c}
\hline
\rule{0pt}{2ex} Stage & MDSNet10 & MDSNet18 & MDSNet34 & MDSNet104 \rule[-1ex]{0pt}{0pt} \\
\hline
\rule{0pt}{3ex} Encoding & \multicolumn{4}{c}{Encoding Block} \rule[-2ex]{0pt}{0pt} \\
\hline
\rule{0pt}{4ex} Conv1 &
MS2Block2(0)  & 
\makecell{MS2Block2(1)} &
\makecell{MS2Block2(2)} &
\makecell{MS2Block2(2)}
\rule[-3ex]{0pt}{0pt} \\
\hline
\rule{0pt}{4ex} Conv2 &
MS2Block2(0)  & 
\makecell{MS2Block2(1)} &
\makecell{MS2Block2(1) \\ MS2Block1(1)} &
\makecell{MS2Block2(3) \\ MS2Block1(3)}
\rule[-3ex]{0pt}{0pt} \\
\hline
\rule{0pt}{4ex} Conv3 &
MS2Block2(0)  & 
\makecell{MS2Block2(1)} &
\makecell{MS2Block2(2) \\ MS2Block1(2)} &
\makecell{MS2Block2(15) \\ MS2Block1(15)}
\rule[-3ex]{0pt}{0pt} \\
\hline
\rule{0pt}{4ex} Conv4 &
MS2Block2(0)  & 
\makecell{MS2Block2(1)} &
\makecell{MS2Block2(2)} &
\makecell{MS2Block2(3) \\ MS2Block1(3)}
\rule[-3ex]{0pt}{0pt} \\
\hline
\rule{0pt}{3ex} Conv5 & \multicolumn{4}{c}{SpikeSPPF} \rule[-2ex]{0pt}{0pt} \\
\hline
\end{tabular}%
}
\label{tab_10}
\end{table}

\section{More Experiments}\label{more_setup}
\subsection{More Implementation Details}
The MDSNet structure is detailed in Table \ref{tab_10}. For SpikeDet-L, we increase the channel dimensions of MDSNet104 by a factor of 1.25 to improve model performance.
On the Gen1 dataset, we employ the zoom-in and zoom-out augmentation strategies from \cite{gehrig_recurrent_2023}. The model is trained for 100 epochs with a batch size of 64.
For URPC 2019, ExDARK, and CrowdHuman datasets, we employ mosaic augmentation. Specifically, following \cite{li_suyolo_2025}, we resize images to 320$\times$320 for URPC 2019, while using 640$\times$640 for the other datasets. All models are trained for 300 epochs with a batch size of 64.

\begin{figure}[!t]
\centering
    \includegraphics[width=.99\columnwidth]{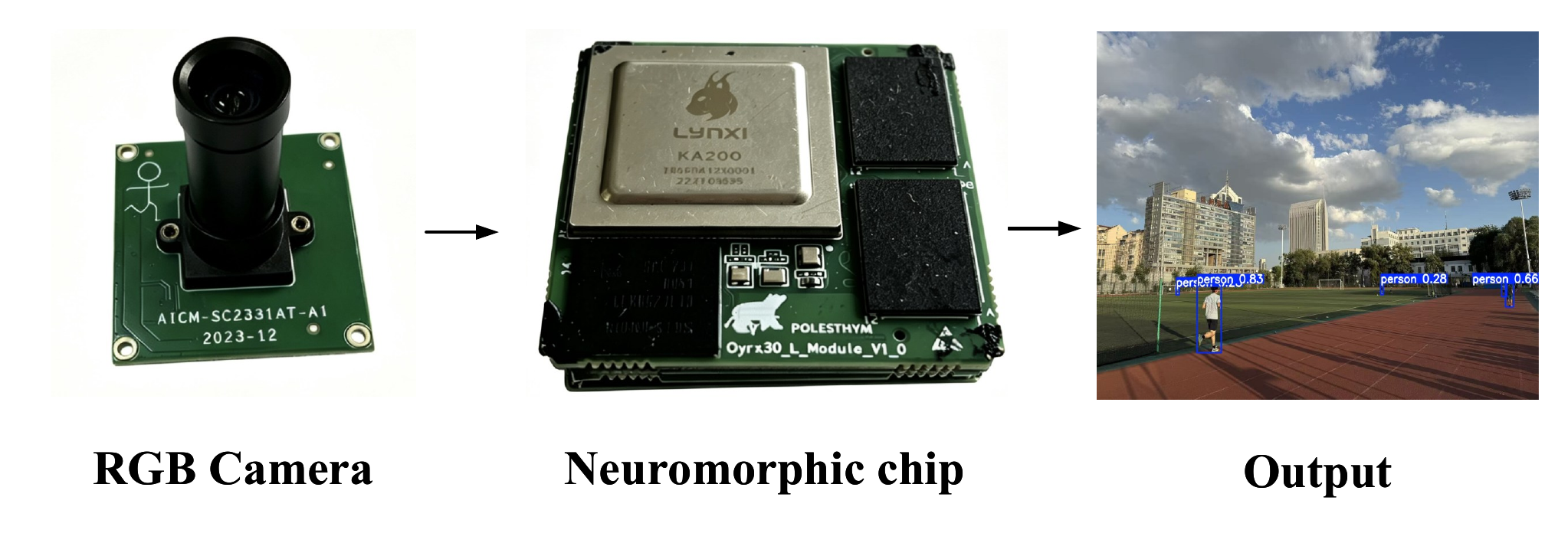}
\caption{\textbf{Execution pipeline of SNNs on neuromorphic chips.}}
\label{fig_neu}
\vspace{-0.2cm}
\end{figure}

For SNN deployment, the LynBIDL framework provided by Lynxi Technologies is adopted for model adaptation without additional restructuring of the inference logic. The model is quantized to FP16 precision. At the macro-core level, the KA200 performs graph compilation with automated compute graph optimization and core resource allocation. As shown in Fig. \ref{fig_neu}, inference is then executed asynchronously on the KA200 APU in a spike-driven manner.
For ANN deployment on the NVIDIA Jetson Xavier NX, TensorRT is employed for inference acceleration with FP16 precision.  Both platforms use an input resolution of 640$\times$640, and the net inference power consumption is measured by subtracting the idle power from the runtime power, ensuring rigorous evaluations.

\subsection{More Dataset Details}
The GEN1 dataset \cite{tournemire_large_2020} represents the initial large-scale collection for object detection using event cameras. It comprises car footage spanning over 39 hours, captured by the GEN1 device with a spatial resolution of 304$\times$240. The dataset includes bounding box annotations for vehicles and pedestrians, provided at rates of 1 to 4Hz. These labels accumulate to a total exceeding 255,000 across the recordings.

The URPC 2019 dataset \cite{wang_ulo_2022} is designed for the 2019 Underwater Robot Picking Contest and comprises 4,707 annotated underwater images across four marine object categories: scallops, starfish, echinus, and holothurian. The dataset splits include 3,767 training images, 695 validation images, and 245 test images. 

The ExDARK dataset \cite{loh2019getting} is designed for low-light object detection, featuring 7,363 images captured under various illumination conditions ranging from extremely dark environments to twilight scenarios. The dataset encompasses 12 object categories: Bicycle, Boat, Bottle, Bus, Car, Cat, Chair, Cup, Dog, Motorbike, People, and Table, with bounding box annotations provided for all instances. In this study, we split the dataset into training, validation, and test sets with a ratio of 6:2:2.

The CrowdHuman dataset \cite{shao2018crowdhuman} focuses on human detection in densely populated scenarios. It comprises approximately 470K annotated human instances across its training and validation sets, with an average density of 22.6 individuals per image and substantial occlusion variations. Annotations include bounding boxes for the head, visible body region, and complete body for each person. The dataset provides 15,000 training images, 4,370 validation images, and 5,000 test images, featuring comprehensive annotations across diverse scenarios.

\subsection{More Benchmark Settings}
\noindent\textbf{Average Precision (AP).} Following \cite{duan_centernet_2023, luo_integervalued_2025}, we adopt AP as the primary detection metric. AP summarizes the precision-recall curve by computing  the area under it. Following the COCO evaluation protocol \cite{lin_microsoft_2014}, AP is averaged over ten IoU thresholds from 0.50 to 0.95 with a step of 0.05. We additionally report AP$_{50}$ and AP$_{75}$, which evaluate detection accuracy at IoU thresholds of 0.50 and 0.75, respectively. To assess scale sensitivity, we further report AP$_s$, AP$_m$, and AP$_l$ for small (area $< 32^2$), medium ($32^2 \leq$ area $< 96^2$), and large (area $\geq 96^2$) objects.

\noindent\textbf{Firing Rate.} The firing rate, which measures neuronal activity, is another critical metric for evaluating SNNs. It is calculated as the average ratio of neuron spikes to the total number of neurons across all time steps. 

\begin{figure}[!t]
\centering
    \includegraphics[width=1.\columnwidth]{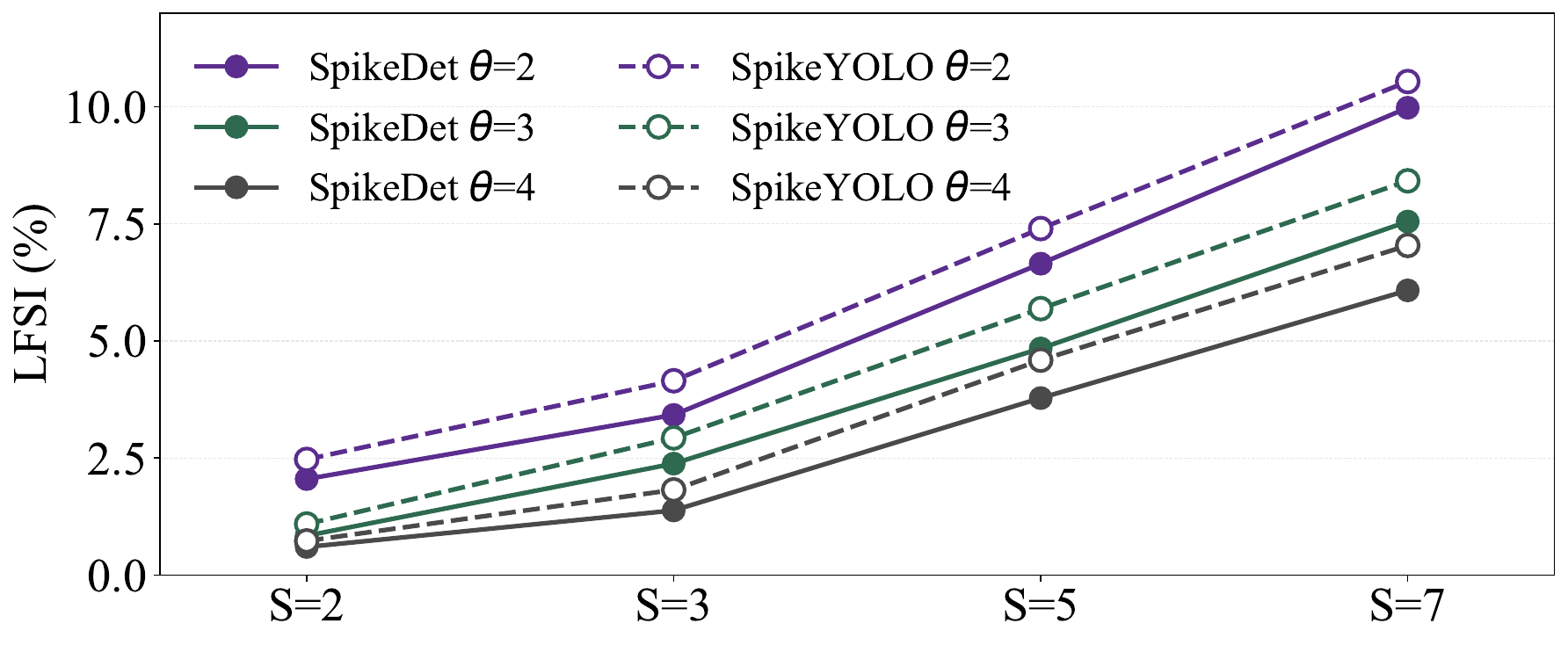}
\caption{Sensitivity analysis of $\theta$ and $S$ in LFSI on SpikeDet and SpikeYOLO.
}
\label{fig_lfsi}
\end{figure}

\begin{figure}[!t]
\centering
    \includegraphics[width=1.\columnwidth]{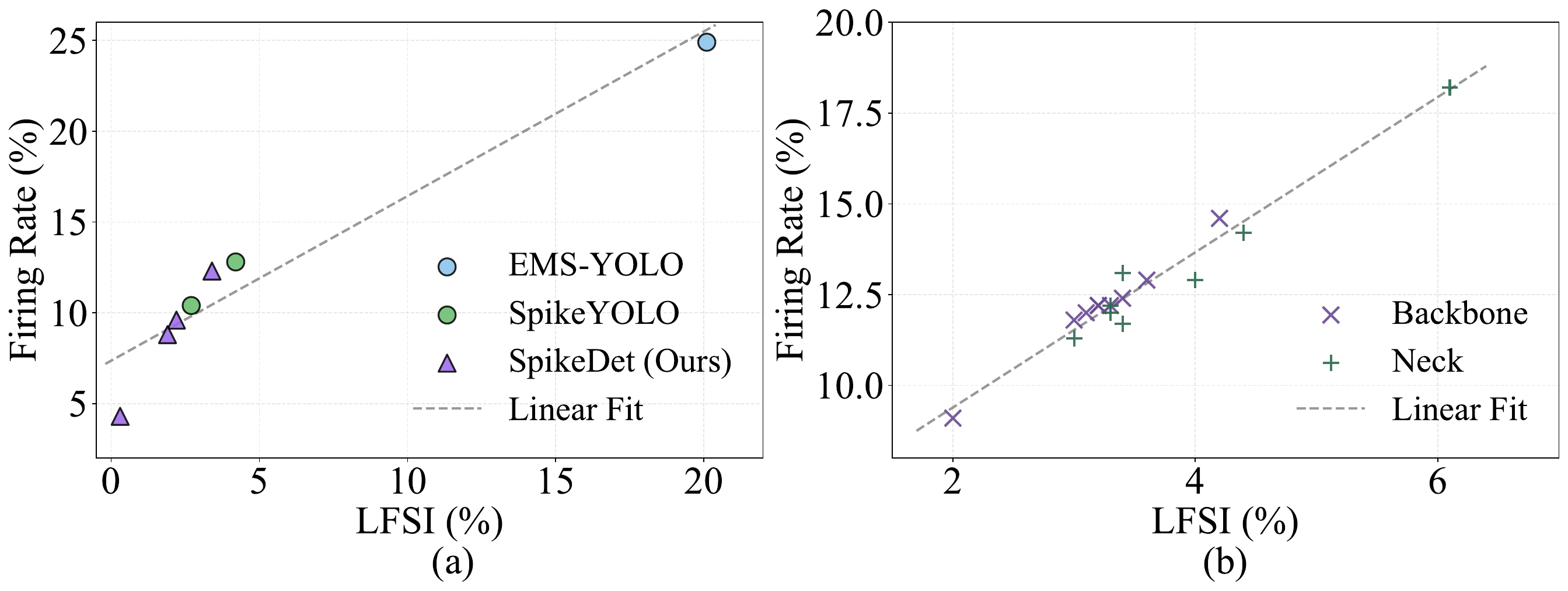}
\caption{\textbf{Correlation analysis of LFSI and firing rate across different SNN-based object detectors.} (a) compares fully converged models, while (b) presents results from backbone and neck ablation studies.
}
\label{fig_lfsi_fr}
\vspace{-0.3cm}
\end{figure}

\subsection{More Analysis on LFSI}
\subsubsection{Sensitivity Analysis of $\theta$ and $S$ in LFSI} 
The parameters $\theta$ and $S$ critically affect LFSI evaluation strictness. Smaller $S$ requires saturated neurons to be more spatially contiguous, while larger $\theta$ demands more neighboring neurons to reach saturation simultaneously, both corresponding to stricter evaluation criteria. 
As shown in Fig.~\ref{fig_lfsi}, overly strict settings ($\theta=4$, $S=2$) drive the LFSI of both models toward zero, reducing inter-model discriminability. Conversely, overly lenient settings ($\theta=2$, $S=7$) introduce excessively large neighborhoods that incorporate spatially non-adjacent saturated neurons, similarly diminishing inter-model discriminability.
Among all configurations, $S=3$ maintains stable inter-model discriminability across different $\theta$ values, and $\theta=2$ exhibits the same property across different $S$ values. Therefore, we adopt $S=3$ and $\theta=2$ as the default parameters for LFSI in subsequent experiments, as this combination ensures evaluation strictness while effectively capturing actual local firing saturation phenomena. 
Furthermore, the LFSI of our method is consistently lower than that of SpikeYOLO \cite{luo_integervalued_2025} across all settings, demonstrating its effectiveness in mitigating the local firing saturation problem.

\subsubsection{Correlation Analysis Between LFSI and Firing Rate}
Firing rate directly measures spiking activity and is a primary determinant of SNN energy consumption. If LFSI reliably captures local firing saturation, it should exhibit a consistent positive relationship with firing rate across model variants. 
As shown in Fig. \ref{fig_lfsi_fr} (a), this is confirmed across fully converged detectors (Pearson $r=0.962$, $p<0.001$), where models with higher local firing saturation tend to maintain elevated overall spiking activity. The same trend appears in our ablation study in Fig. \ref{fig_lfsi_fr} (b). LFSI thus effectively reflects model spiking activity levels, serving as a reliable quantitative metric for evaluating the energy efficiency of SNN-based detectors.

\begin{table}[tbp]
\centering
\caption{Impact of Different Time Steps}
\setlength{\tabcolsep}{3.5pt}
\begin{tabular}{lccccccc}
\hline
Model & T$\times$D & AP & $\text{AP}_{50}$ & \makecell[l]{Param \\ (M)} & \makecell[l]{Firing \\ Rate(\%)} &\makecell{LFSI\\(\%)} & \makecell[c]{Energy\\(mJ)} \\
\hline
\multirow{3}{*}{SpikeDet} & 1$\times$4 & 44.8 & 61.2 & 22.0 & 11.8 & 3.0 & 19.0 \\
& 1$\times$6 & 45.6 & 62.3 & 22.0 & 8.8 & 1.0 & 20.9 \\
& 1$\times$8 & 46.1 & 62.7 & 22.0 & 7.7 & 0.5 & 23.3 \\
\hline
\end{tabular}
\label{tab_time}
\end{table}

\begin{table}[tp]
\centering
\caption{\textbf{Comparison of different Fusion Blocks.}
SF-Block- and SF-Block+ refer to removing one MS-Block from SF-Block and adding an additional MS-Block, respectively.}
\begin{tabular}{lcccccc}
\hline
\makecell[l]{Fusion \\ Block} & AP & $\text{AP}_{50}$ & \makecell{Param \\ (M)} & \makecell{Firing \\ Rate(\%)} &\makecell{LFSI\\(\%)} & \makecell{Energy\\(mJ)} \\
\hline
7$\times$7Conv & 41.7 & 58.5 & 22.6 & 11.6 & 4.0 & 12.5 \\
MS-Block & 43.6 & 60.7 & 18.7 & 13.3 & 3.6 & 16.3 \\
SF-Block- & 43.8 & 60.9 & 20.3 & 12.0 & 3.2 & 17.8 \\
SF-Block+ & 44.6 &  61.2 & 27.9 & 11.7 & 3.0 & 22.7 \\
\textbf{SF-Block} & \textbf{44.8} & \textbf{61.2} & \textbf{22.0} & \textbf{11.8} & \textbf{3.0} & \textbf{19.0} \\
\hline
\end{tabular}
\label{tab_fusion_block}
\end{table}

\begin{table}[H]
\centering
\caption{\textbf{Ablation study on the downsampling operations within MDS.}
Pre. and Post. refer to placing MaxPool before and after the spiking neuron, respectively, and Stride indicates whether a parallel stride path is included.}
\setlength{\tabcolsep}{3.5pt}
\begin{tabular}{ccccccccc}
\hline
Pre. & Post. & Stride & AP & $\text{AP}_{50}$ & \makecell{Param \\ (M)} & \makecell{Firing \\ Rate(\%)} & \makecell{LFSI \\ (\%)} & \makecell{Energy \\ (mJ)} \\
\hline
$\checkmark$ & & & 44.5 & 61.0 & 22.0 & 12.3 & 3.1 & 19.6 \\
& $\checkmark$ & & 44.5 & 60.9 & 22.0 & 12.3 & 3.1 & 19.7 \\
$\checkmark$ & & $\checkmark$ & 44.6 & 61.1 & 22.0 & 12.1 & 3.1 & 19.4 \\
\textbf{} & $\boldsymbol{\checkmark}$ & $\boldsymbol{\checkmark}$ & \textbf{44.8} & \textbf{61.2} & \textbf{22.0} & \textbf{11.8} & \textbf{3.0} & \textbf{19.0} \\
\hline
\end{tabular}
\label{tab_mds_down}
\end{table}

\subsection{More Ablation Studies}\label{more_ablation}
\subsubsection{Ablations on Time Step}
Increasing time steps enhances the encoding capacity of SNN-based models, reduces quantization error, and thereby indirectly alleviates local firing saturation. To verify this, we conduct ablation experiments on time steps. Since adjusting T is inefficient~\cite{luo_integervalued_2025}, we control the overall time steps by adjusting D instead. As shown in Table~\ref{tab_time}, consistent with our analysis, increasing D effectively reduces firing rates and LFSI, but at the cost of higher energy consumption.

\subsubsection{Ablations on SF-Block}
We replace the SF-Block in SMFM with the simplest single convolution to demonstrate its effectiveness.
As shown in Table \ref{tab_fusion_block}, SF-Block significantly outperforms it in AP, LFSI, and firing rate, with only a slight increase in energy consumption.
Meanwhile, we conduct ablation on the two sub-blocks of SF-Block. Rows 2 and 3 in Table \ref{tab_fusion_block} retain only the MS-Block and the first sub-block, respectively, while row 4 extends the MS-Block to two instances. The results demonstrate that the proposed configuration effectively addresses the local firing saturation problem while achieving a better trade-off between performance and power consumption.

\subsubsection{Ablations on Downsampling Operations in MDS}
As shown in Table \ref{tab_mds_down}, we first conduct ablation using only MaxPool for downsampling. Consistent with \cite{zhou2023enhancing}, placing MaxPool before I-LIF yields better performance due to more accurate gradient information. However, since I-LIF provides stronger spatial representation than LIF, the performance gap is not significant. Rows 3 and 4 further validate the effectiveness of introducing fixed stride downsampling. With the stride path included, placing the downsampling operations after I-LIF effectively alleviates local firing saturation and improves performance, consistent with our analysis.

\begin{figure}[t]
\centering
    \includegraphics[width=.99\columnwidth]{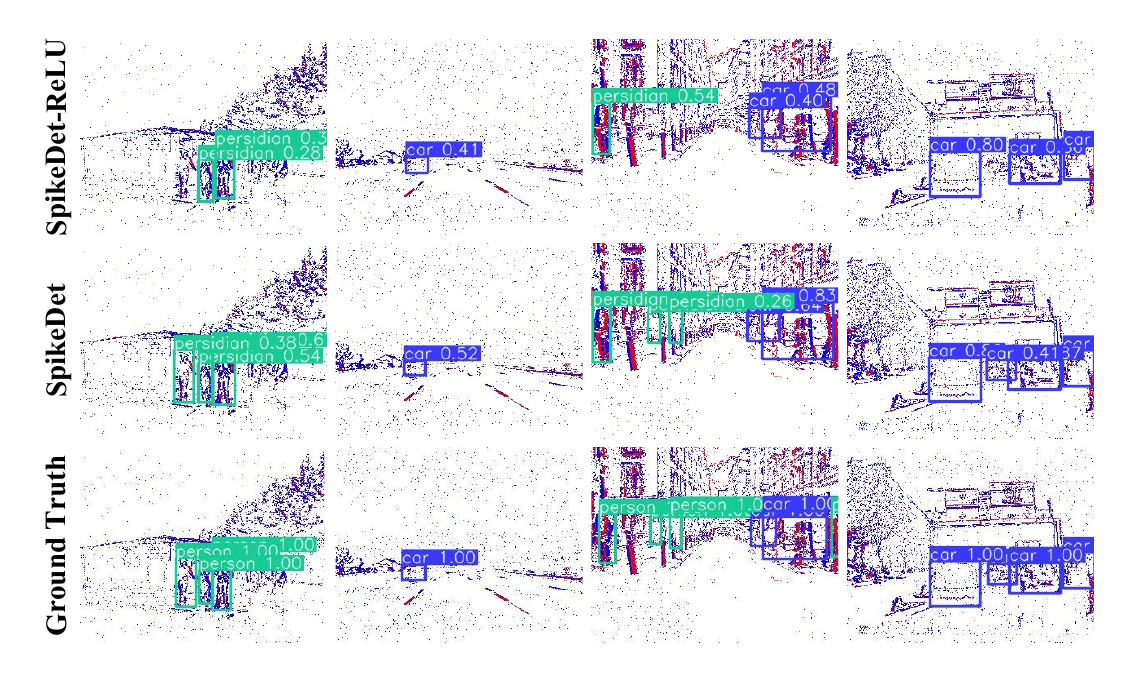}
\caption{\textbf{Detection results on the GEN1 dataset.}}
\label{fig_gen1}
\vspace{-0.2cm}
\end{figure}

\section{More Visualization}
In this section, we present the visualization results of our model on the GEN 1 dataset in Fig. \ref{fig_gen1}.

\bibliographystyle{IEEEtran}
\bibliography{references}

\end{document}